\documentclass{article}
\usepackage[numbers]{natbib}
\usepackage {arxiv}

\usepackage{floatrow}
\usepackage{hyperref}
\usepackage{amsmath}
\usepackage{amssymb}
\usepackage{mathtools}
\usepackage{amsthm}
\usepackage{amsmath,amssymb,amsfonts}
\usepackage{algorithm}
\usepackage{algpseudocode}
\usepackage{textcomp}
\usepackage{xcolor}
\usepackage{float}
\usepackage{booktabs}
\usepackage{multirow}
\usepackage{enumitem}
\usepackage{setspace}
\usepackage[]{graphicx}
\usepackage{array}
\usepackage{amsmath}
\usepackage{amssymb}
\usepackage{epsfig}
\usepackage{epstopdf}
\usepackage{comment}
\usepackage{tabularx}
\usepackage{threeparttable}
\usepackage{adjustbox}
\usepackage{rotating}
\usepackage{dsfont}
\usepackage{hyperref} 
\usepackage{lineno}
\usepackage{soul}
\usepackage{tabularx}
\usepackage{multirow}
\usepackage{booktabs}
\usepackage{subcaption}
\usepackage{pdflscape}
\usepackage{adjustbox}
\usepackage{array}
\usepackage{makecell}
\RequirePackage{fix-cm}
\usepackage{amsmath}
\usepackage{tikz}
\usepackage{booktabs}
\usepackage{adjustbox}
\usepackage{subcaption}
\usepackage{float}
\usetikzlibrary{arrows.meta,positioning,calc}

\title{\LARGE \bf
Copula Adapted Directed Acyclic Graph for Cluster Representation of  Biomedical Data}

\author{Heranga K. Rathnasekara ~ ~ Norou Diawara \\
       Department of Mathematics and Statistics\\
Old Dominion University, \\
Norfolk, VA 23529, USA\\
\texttt {\{hrath001, ndiawara\}@odu.edu }
\and 
{\bf Manar D. Samad}\\
Department of Computer Science\\
North Carolina Agricultural and Technical State University\\
Greensboro, NC 27411, USA\\
\texttt{mdsamad@ncat.edu} \\
}

\begin{document}

\maketitle

\begin{abstract}

Diagnostic errors and mislabeling are common in biomedicine, which compromise the reliability of predictive models and data-driven outcomes. Stratifying unlabeled biomedical data based on complex relationships between features eliminates the need for data labels and overcomes the limitations of supervised learning. Traditional clustering methods assume restrictive data distributions, making them suboptimal for capturing complex dependencies in high-dimensional biomedical data. This paper introduces a novel cluster-friendly data presentation framework that integrates the non-Gaussian and non-linear feature dependence of copula models with an ensemble of causal structure discovery (CSD) methods based on Directed Acyclic Graphs (DAGs). While copulas model flexible multivariate distributions by relaxing assumptions related to multivariate normality, linear dependence, and symmetric relationships, an ensemble of DAG-based CSD methods identifies stable causal relationships between features. When clustered using K-means, the new data representation obtained by the proposed copula-adapted DAG (CopDAG) ranks first among the 12 methods in normalized clustering accuracy and adjusted Rand index across 16 biomedical datasets. Our CopDAG method predicts ground-truth class labels directly from feature relationships without data annotations and supervised learning, while also providing cluster visualizations and explainable causal structures of the biomedical data features.

\end{abstract}


\section{Introduction}
In many medical and scientific research contexts, acquiring reliable outcome or response features is challenging, expensive, and labor-intensive \citep{Arazo2019}. Diagnostic codes documented in medical records do not necessarily ensure that the associated diagnosis label is accurate. Diagnostic mistakes—including missed, delayed, or incorrect diagnoses—remain a significant threat to patient safety. Makary and Daniel \citep{MakaryDaniel2016} estimated that medical errors, broadly defined, may constitute the third leading cause of death in the United States. The National Academies of Sciences, Engineering and Medicine further concluded that most individuals are likely to experience at least one diagnostic error during their lifetime \citep{NASEM2015Diagnosis}. In biomedical research, mislabeled observations compromise the validity of subsequent statistical modeling and predictive analytics. Inaccurate labels can result from diagnostic errors, systemic or institutional biases, unequal class distributions, or sampling distortions. Such data severely limit the ability of predictive models to represent the underlying population accurately \citep{Arazo2019}.

In this context, unsupervised clustering methods provide a complementary perspective because response features are not used when learning latent groups in the data. Instead, clustering methods leverage internal structures and relationships between observed features to identify naturally occurring subgroups \citep{koutroumbas2008pattern}. Therefore, clustering solutions alleviate most of the aforementioned concerns related to data labels and can further discover new subgroups or phenotypes beyond known clinical stratification. However, traditional clustering methods are applied directly to raw observations, relying on Euclidean distances and restrictive distributional assumptions, and overlooking complex nonlinear, asymmetric, and conditional relationships among features \citep{koutroumbas2008pattern}. Furthermore, the ability of traditional clustering methods to recover true class labels from unlabeled data remains underexplored and challenging. This paper introduces a novel \emph{cluster-friendly} data representation method that combines the strength of directed acyclic graphs (DAGs) in learning causal and directional dependencies between features with multivariate copula models built on non-linear and conditional dependencies between variables. Grounded in statistical theory, a computational framework called copula adapted DAG or CopDAG, which integrates copulas with DAGs, transforms raw data into {cluster-friendly} representations to recover ground-truth labels in biomedical datasets, while simultaneously offering interpretable feature relationships and clear visualizations of the resulting clusters.

The remainder of the paper is organized as follows. Section~\ref{relatedwork} highlights recent work relevant to our study. Section~\ref{sec:background} presents the background and notation. Section~\ref{sec:methods} introduces the proposed framework. Section~\ref{sec:experiments} describes the real-data used for this study, while Section~\ref{sec:results} presents the corresponding results. The subsequent sections provide discussion and conclusions.

\subsection{Related work}\label{relatedwork}

Recent biomedical studies have used copula models to characterize complex dependence among biomarkers \citep{ma2023flexible, liu2023decomposition,10.3389/fendo.2024.1291895} and for synthetic data generation \citep{guo2024generation}. Ma et al. \citep{ma2023flexible} introduced a copula-based framework for integrating correlated single-cell multi-omics datasets that captures dependencies separately from marginal distributions and incorporates biomedical characteristics. A latent mixed Gaussian copula model was used to quantify associations among binary, ordinal, continuous, and truncated biomedical features and to identify common and group-specific variation \citep{liu2023decomposition}. In genomic applications, copula models have been used to characterize changes in gene-pair dependence under different biological conditions \citep{ray2020codc}. In clinical applications, copula-based analysis revealed significant dependence among diabetes-related biomarkers \citep{10.3389/fendo.2024.1291895}, while flexible copula regression was applied to capture asymmetric relationships among markers of glycemic control ~\citep{EspasandinDominguez2019}. Most recently, vine copulas have been shown to meaningfully capture the dependencies in features derived from Electronic Health Records (EHR) by forming hierarchical tree structures shaped by vines \citep{samad2026miningelectronichealthrecords}. Copula models have been increasingly used in biomedical studies because they can characterize complex dependence separately from the marginal distributions. However, most existing applications focus on estimating and interpreting dependence rather than providing a clear framework for translating feature dependence into \emph{cluster-friendly} data representations.

To incorporate dependence directly into clustering, researchers have developed copula-based approaches. For example, A copula-based clustering method (CoClust)  \citep{CoClust} groups multivariate observations according to their underlying dependence structure, while Afrin et al. \citep{afrin2020directionally} demonstrated the relevance of directional dependence for biomedical clustering by showing that ignoring directional relationships among multiple omics data types can reduce clustering performance. However, these approaches embed dependence within a specific clustering model rather than developing a general dependence-guided representation that can be used with different clustering algorithms. The limitations of copula methods can be addressed by incorporating causal and directional structure of features. Causal relationships between features, modeled via DAGs can provide complementary information about data characteristics. DAGs provide a complementary approach for representing conditional and potentially directional relationships among features and have been widely used in biomedical studies. Yang et al. developed a model-free approach to estimate DAG skeletons and identified nonlinear relationships between gene pairs that Gaussian graphical-model methods did not capture \citep{yang2021modelfree}. However, the approach focuses on estimating the DAG skeleton rather than the directions of the identified relationships. More recently, a DAG-based phenotype-network method was developed to estimate directed relationships among cardiovascular-related proteins associated with Alzheimer’s disease \citep{zilinskas2024inferring}. However, this approach assumes a Gaussian linear structural model, which may limit the representation of complex nonlinear relationships.

Thus, copula-based graphical models provide a flexible alternative by relaxing distributional assumptions and allowing nonlinear and non-Gaussian data dependence to be modeled within a graphical structure. Although these approaches combine the strengths of copula and graphical models, existing studies have primarily focused on network inference and interpretation rather than clustering. It is also unknown how conditional feature relationships can transform data to facilitate clustering.

\section{Preliminaries}
\label{sec:background}

This section presents the theoretical foundations of copulas and directed acyclic graphs used to model causal relationships.

\subsection{Copula Theory and Sklar's Theorem}
A copula is a multivariate distribution function with uniform marginals that captures the dependence structure of a random vector independently of its marginal distributions. The foundational result is Sklar's (1959) theorem \citep{Sklar1959}, which states that for any joint distribution \(F\) with marginals \(F_1,...,F_p\), there exists a copula \(C\) such that
\[
F(x_1,\dots,x_p) = C\big(F_1(x_1),\dots,F_p(x_p)\big),
\]
where $C:[0,1]^d \to [0,1]$ is unique when the marginals are continuous. The corresponding joint density factorizes as
\begin{equation}
f(x_1,\dots,x_p)
= c\big(u_1,\dots,u_p\big)\,\prod_{j=1}^p f_j(x_j),
\label{eq:sklar-density}
\end{equation}
where $u_j = F_j(x_j)$, $f_j$ is the marginal density of $X_j$, and 
\[
c(u_1,\dots,u_p)
= \frac{\partial^p}{\partial u_1 \cdots \partial u_p}
  C(u_1,\dots,u_p)
\]
is the copula density. This factorization separates the marginal behavior from their dependence structure. The probability integral transformation $U_j = F_j(X_j)$ produces uniform features on $[0,1]$, while the copula captures the dependence among them.
The marginal representation of each feature in the copula provides two-fold benefits. First, marginal distributions capture conditional relationships between features without any distribution assumptions. Second, all marginals are uniform distributions, which is critical for modeling biomedical data with varying distributions, scales, and degrees of skewness. The conditional dependence in copulas is captured using different families of copulas without requiring the raw data to make any distribution assumption.

Different forms of dependence from Eq. \ref{eq:sklar-density} lead to different copula families. Elliptical copulas, such as the Gaussian and Student-$t$ copulas, describe symmetric dependence structures. Archimedean families provide additional flexibility. For example, Clayton and Gumbel copulas capture lower- and upper-tail dependence respectively, while Frank copulas allow flexible dependence without tail dependence. These properties make copulas well suited to capturing complex relationships among biomarkers, particularly when simple linear correlations do not adequately represent these dependencies.

\subsection{Vine Copulas}

The copula families described above can model multivariate dependence through a single copula structure. However, this approach can become restrictive as the number of features increases, particularly when different feature pairs exhibit different forms of dependence. Vine copulas provide greater flexibility by decomposing multivariate dependence into a sequence of conditional bivariate copulas, or pair-copulas, organized through linked trees \citep{BedfordCooke2001,AasCzadoFrigessiBakken2009}. This flexibility allows different copula families to be selected for different feature pairs. Among vine structures, the regular vine (R-vine) provides a flexible construction because it does not impose the specific structural forms of C-vines or D-vines.

Let $\mathbf{U} = (U_1,\dots,U_p)$ denote the probability integral transforms of the features.  
A vine copula factorizes the copula density as
\begin{equation*}
c(u_1,\dots,u_p) 
= \prod_{m=1}^{p-1} \prod_{e\in E_m} 
    c_{j_e, k_e; D_e}
    \big(u_{j_e \mid D_e}, u_{k_e \mid D_e}\big),
\end{equation*}
where $E_m$ denotes the edge set of tree $m$, $j_e$ and $k_e$ denote the features associated with edge $e$, and each factor $c_{j_e,k_e;D_e}$  
that may be conditioned on the set 
$D_e$.

In this study, five candidate families--Gaussian, Student-(t), Clayton, Gumbel, and Frank--were considered, where the best fit family for each bivariate copula is selected using the Akaike information criterion (AIC).

\subsection{DAG-Based Causal Structure}
\label{sec:DAG definition}
A directed acyclic graph (DAG) is represented as ${\mathcal{G}}=(V,E)$, where 
$V=\{1,\ldots,p\}$ denotes the set of nodes associated with features 
$X_1,\ldots,X_p$, and $E \subseteq V \times V$ denotes the set of directed 
edges that contain no directed cycles \citep{spirtes2000causation}. 
An edge $i \rightarrow j$ indicates relationships of the form
\[
X_i \rightarrow X_j \quad \Longrightarrow \quad X_i \text{ is a parent of } X_j.
\]
DAG-based causal structure learning algorithms (e.g., 
Hill-Climbing \citep{selman2006hill}, PC \citep{spirtes2000causation}) identify the minimal set of edges needed to reproduce the conditional 
independence structure of the data. DAG-based causality structures can facilitate data clustering in three ways. First, a DAG determines the parent set belonging to each feature and the Markov blanket. The Markov blanket of a node consists of its parents, children, and other parents of its children, identifying conditional relationships among the most directly associated features. Second, under standard assumptions, DAG edges may admit causal interpretation, enabling mechanistic exploration of pathways among interrelated biomarkers. Third, a DAG-based causal framework yields a sparse graphical structure by removing unnecessary edges, simplifying subsequent dependence modeling, and helping focus the analysis on the most relevant biomarker relationships.

Let $X = (X_1, \dots, X_p)$ be a set of random features whose dependence structure is represented by a DAG.  For each node \(X_j\): 
\begin{equation*}
X_j=f_j(pa(X_j))+\epsilon_j
\end{equation*}

\noindent
where $\mathrm{pa}(j)$ denote the set of parent nodes of $X_j$ in the DAG and \(\epsilon_j\) = error term. The joint density factorizes according to the DAG as

\begin{equation}
\label{eq:joint density factorizes according to the DAG}
f(x_1, \dots, x_p)
=
\prod_{j=1}^{p}
f(x_j \mid x_{\mathrm{pa}(j)}).
\end{equation}

Such factorization in Eq. \ref{eq:joint density factorizes according to the DAG} specifies the conditional structure of the joint distribution but does not impose a specific form for the dependence between features. 
\subsection{R-vine Graphical Representation}
\label{sec:Rvine_definition}

Unlike a DAG, an R-vine is represented by a sequence of linked undirected trees. Following \citep{dissmann2013selecting}, let
\[ \mathcal{T}=(T_1,\ldots,T_{p-1}) \] denote an R-vine on $p$ features. The first tree is defined as \[ T_1=(V,E_1^{'}), \] where $V=\{1,\ldots,p\}$ denotes the set of nodes corresponding to features $X_1,\ldots,X_p$, and $E_1^{'}$ denotes the set of undirected edges. For $m=2,\ldots,p-1$, the nodes of tree $T_m$ are the edges of the preceding tree, such that \[ T_m=(E_{m-1}^{'},E_m^{'}). \]

\subsection{Copula-DAG Formulation}

Copula and DAG play complementary roles in the proposed framework. Before moving to the proposed framework, we first present the mathematical formulation of the copula and DAG to make the proposed CopDAG framework easier to understand.

Consider a features $X_d$, where $d \in \{1,...,p\}$. 
If the parents of $X_d$ are, $\mathrm{pa}(X_d) = (X_{k_1}, \dots, X_{k_m})$, then the joint distribution linking $X_d$ and its parents can be written using a copula density as
\begin{equation*}
f(x_d, x_{k_1}, \dots, x_{k_m})
=
c_d
\Big(
u_d, u_{k_1}, \dots, u_{k_m}
\Big)
\Big(\prod_{r=1}^{m} f_{k_r}(x_{k_r})\Big)f_d(x_d),
\end{equation*}

where \(u_d = F_d(x_d),~
u_{k_r} = F_{k_r}(x_{k_r})\), and $c_d(\cdot)$ is the copula density.

Similarly, if the children of $X_d$ are 
$\mathrm{ch}(X_d) = (X_{h_1}, \dots, X_{h_q})$, then the joint density linking $X_d$ and its children is
\begin{equation*}
f(x_{h_1}, \dots, x_{h_q},x_d)
=
c_d
\Big(
u_{h_1}, \dots, u_{h_q},u_d
\Big)
\Big(\prod_{r=1}^{q} f_{h_r}(x_{h_r})\Big)f_d(x_d),
\end{equation*}
where \(u_{h_r} = F_{h_r}(x_{h_r}), \text{and} ~u_d = F_d(x_d)\).
Considering both the parent- and children-based formulation, DAG factorization is defined in terms of parent sets. Thus, for $X_d$, the corresponding conditional distribution is
\begin{equation*}
f(x_d \mid x_{\mathrm{pa}(X_d)}; \theta_d, \psi_d),
\end{equation*}
where $\theta_d$ and $\psi_d$ denote the parameters of the conditional distribution of $X_d$ and the parameters of its dependence on the parent features, respectively.

As an illustrative example, suppose the $X_d$ has parents \((X_1,...,X_{d-1})\) and children \((X_{d+1},...,X_{p})\). 
\begin{center}
\begin{minipage}[c]{0.35\textwidth}
\centering
\[
\begin{array}{c@{\;}c@{\;}c}
X_1     & \to    & X_d, \\[-3pt]
        & \vdots &      \\[-3pt]
X_{d-1} & \to    & X_d, \\[4pt]
X_d     & \to    & X_{d+1}, \\[-3pt]
        & \vdots &          \\[-3pt]
X_d     & \to    & X_p.
\end{array}
\]
\end{minipage}
\hfill
\begin{minipage}[c]{0.55\textwidth}
\centering
\begin{tikzpicture}[
    >=Stealth,
    every node/.style={inner sep=2pt}
]
  \node (X1)    at (-2,1)  {$X_1$};
  \node (dots1) at (-2,0)    {$\vdots$};
  \node (Xd-1)    at (-2,-1) {$X_{d-1}$};

  \node (Xd) at (0,0) {$X_d$};

  \node (Xd+1)   at (2,1)  {$X_{d+1}$};
  \node (dots2) at (2,0)    {$\vdots$};
  \node (Xp)    at (2,-1) {$X_p$};

  \draw[->] (X1) -- (Xd);
  \draw[->] (Xd-1) -- (Xd);
  \draw[->] (Xd) -- (Xd+1);
  \draw[->] (Xd) -- (Xp);
\end{tikzpicture}
\end{minipage}
\end{center}
Note that some features may connect indirectly through longer paths, such as $X_1 \to X_2 \to X_d$, where $X_2$ is a direct parent and $X_1$ is an ancestor of $X_d$.

\begin{figure}[t]
\centering
\resizebox{\textwidth}{!}{%
\input{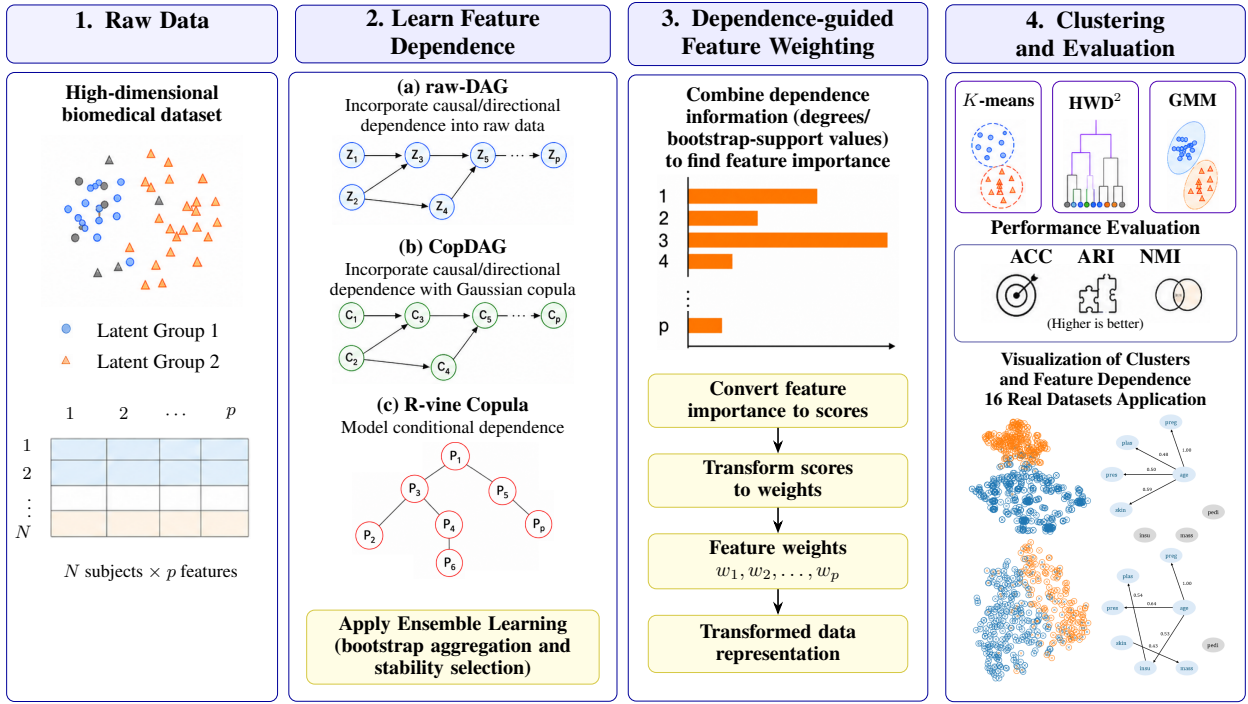}
}
\caption{Computational steps, evaluation, and outcomes of the proposed copula-adapted directed acyclic graph modeling for cluster-friendly data representation}
\label{flowchart1}
\end{figure}

\section{Proposed Framework}
\label{sec:methods}

We propose a Gaussian copula-adapted DAG framework (CopDAG) and compare it with three baseline feature representations. The first baseline uses an ensemble of DAGs learned from raw data (raw-DAG). The second model is R-vine, which is an ensemble of R-vine copulas. In this model, it aggregates the estimated feature-dependence information from each individual model to produce the ensemble model. The third uses the original unweighted data (raw data). Figure~\ref{flowchart1} illustrates the overall computational framework.

\subsection{Preliminary Data Transformation}

Let $X=(X_1,...X_p)\in \Re^{N \times p}$ denote the data matrix comprising $p$ features vectors. Individual features are standardized by the corresponding mean and standard deviation into $Z=(Z_1,..., Z_p)$. Next, $n$ observations are sampled with replacement from $Z$ to form $Z^{(b)}, b=1,...,B$, where, $B$ is the total number of bootstrap iterations and $n \le N$ denotes the bootstrap sample size. Each $Z^{(b)}$ is then transformed into pseudo-observations $P^{(b)}$. The pseudo-observations were used directly for the R-vine method and were further transformed into Gaussian copula scores ($C^{(b)}$) for the CopDAG method. For feature $j$, the pseudo-observations ($P_j$) and Gaussian-copula ($C_j$) scores were defined as
\begin{equation}
\label{equation_pseudo_transform}
P_j^{(b)}
=
\frac{\operatorname{rank}(Z_j^{(b)})-0.5}{n},
\qquad j=1,\ldots,p,
\end{equation}
and
\begin{equation}
\label{equation_gaussian_transform}
C_j^{(b)}=
\Phi^{-1}\left(P_j^{(b)}\right),
\qquad j=1,\ldots,p,
\end{equation}
where $\Phi^{-1}(\cdot)$ is the inverse standard normal cumulative distribution function.

\subsection{Dependence Structure Learning and Ensemble Aggregation}

For raw-DAG and CopDAG approaches, $L=7$  graph-based causality algorithms implemented in the \texttt{bnlearn} package in R were used \citep{citebnlearn}. The selected algorithms were Hill-Climbing (HC) \citep{selman2006hill} with the Bayesian Information Criterion (BIC), tabu search \citep{KITSON2025109522tabu} with BIC, the PC-stable algorithm \citep{spirtes2000causation}, HC with AIC, the Max-Min HC algorithm \citep{Tsamardinos2006}, Incremental Association Markov Blanket (IAMB) \citep{tsamardinos2003algorithms}, and Fast-IAMB.  These methods represent score-based, constraint-based, and hybrid structure-learning approaches. When appropriate, blacklist constraints were incorporated based on prior knowledge or study assumptions to prevent selected arc directions from being learned during DAG learning. To construct the raw-DAG and CopDAG, the $L$ DAGs ($\mathcal{G}=(V,E)$) were applied to $Z^{(b)}$ and $C^{(b)}$, respectively, with

\begin{equation}
\begin{aligned}
\widehat{\mathcal{G}}_{Z,\ell}^{(b)}
    &= \left(V,\widehat{E}_{\ell,Z}^{(b)}\right),
    && \ell=1,\ldots,L, ~~~~b=1,...,B
\end{aligned}
\label{eq:bootstrap_dags_raw}
\end{equation}
and
\begin{equation}
\begin{aligned}
\widehat{\mathcal{G}}_{C,\ell}^{(b)}
    &= \left(V,\widehat{E}_{\ell,C}^{(b)}\right),
    && \ell=1,\ldots,L, ~~~~b=1,...,B
\end{aligned}
\label{eq:bootstrap_dags_gc}
\end{equation}

Within each bootstrap sample, we aggregated the seven learned DAGs using a majority-voting procedure. An arc was retained in the bootstrap-level consensus graph if it appeared in more than ${L}/{2}$ of the learned structures. 

\begin{equation*}
\begin{aligned}
\widehat{{E}}_{Z}^{(b)}
&=
\left\{
(j\rightarrow k):
\sum_{\ell=1}^{L}
I\left[(j\rightarrow k)\in\widehat{E}_{\ell,Z}^{(b)}\right]
> \frac{L}{2}
\right\},\\ \mbox{and} \quad
\widehat{{E}}_{C}^{(b)}
&=
\left\{
(j\rightarrow k):
\sum_{\ell=1}^{L}
I\left[(j\rightarrow k)\in\widehat{E}_{\ell,C}^{(b)}\right]
> \frac{L}{2}
\right\}.
\end{aligned}
\label{eq:bootstrap_consensus_dags_with_majority_voting_edge}
\end{equation*}

The corresponding DAGs are then defined as
\begin{equation}
\begin{aligned}
\widehat{\mathcal{G}}_{Z}^{(b)}=(V,\widehat{{E}}_{Z}^{(b)}), \quad \mbox{and}~~~~~
\widehat{\mathcal{G}}_{C}^{(b)}=(V,\widehat{{E}}_{C}^{(b)})
\end{aligned}
\label{eq:bootstrap_consensus_dags_with_majority_voting}
\end{equation}

For the R-vine approach, an R-vine copula model was fitted to $P^{(b)}$ to obtain the dependence-learning undirected structure $\widehat{\mathcal{T}}_{R}^{(b)}$, with 
\begin{equation}
\label{rvine}
\widehat{\mathcal{T}}_{R}^{(b)}
    =
    \left(V,\widehat{E'}_{R}^{(b)}\right),
\end{equation}
where, $
\widehat {E'}_{R}^{(b)}
=\left((j,k),\left|\hat\tau_{jk}^{(b)}\right|\right)$, \text{ $\hat \tau_{jk}^{(b)}$ denotes the dependence strength (Kendall's}
$\tau$) associated with edge $(j,k)$ in bootstrap iteration $b$. We used the VineCopula R package for this implementation.

After $B$ bootstrap iterations, the three methods were aggregated across bootstrap samples. Edges with bootstrap support greater than or equal to a predefined stability threshold $\gamma$ were retained to construct the consensus graphs as follows:

\begin{equation}
\label{final_consensus_graphs_all_ensemble_1}
\begin{aligned}
\widehat{\mathcal{G}}_{Z}
&=
\left(V,\widehat{E}_{Z}\right),
&
\widehat{E}_{Z}
&=
\frac{1}{B}
\sum_{b=1}^{B}
\mathbb{I}
\left[
(j\rightarrow k)\in\widehat{E}_{Z}^{(b)}
\right]
\geq \gamma, ~~~~~~\text{raw-DAG}
\end{aligned}
\end{equation}
\begin{equation}
\label{final_consensus_graphs_all_ensemble_2}
\begin{aligned}
\widehat{\mathcal{G}}_{C}
&=
\left(V,\widehat{E}_{C}\right),
&
\widehat{E}_{C}
&=
\frac{1}{B}
\sum_{b=1}^{B}
\mathbb{I}
\left[
(j\rightarrow k)\in\widehat{E}_{C}^{(b)}
\right]
\geq \gamma, ~~~~~~\text{CopDAG}
\end{aligned}
\end{equation}
\begin{equation}
\label{final_consensus_graphs_all_ensemble_3}
\begin{aligned}
\widehat{\mathcal{T}}_{R}
&=
\left(V,\widehat{E'}\right),
&
\widehat{E'}_{R}
&=
\frac{1}{B}
\sum_{b=1}^{B}
\mathbb{I}
\left[
(j,k)\in\widehat{E'}_{R}^{(b)}
\right]
\geq \gamma. ~~~~~~~~~~\text{R-vine}
\end{aligned}
\end{equation}

Then, node-level connectivity measures were extracted from each consensus DAG. We introduce the DAG with node-level connectivity representation with $\widehat D$, which denotes the collection of in- and out-connectivity degrees for all features: 
\[
\widehat{\mathcal G}
=
\left(V,\widehat E,\widehat D\right),
\]

where \(\widehat D\) contains the in-degree and out-degree connectivity for each feature. Specifically, for raw-DAG:
\begin{equation}
\label{indegree_and_outdegree_1}
\widehat D_Z
=
\left\{
\left(
\widehat D_{\mathrm{in}}(Z_j),
\widehat D_{\mathrm{out}}(Z_j)
\right):
j=1,\ldots,p
\right\},
\end{equation}

where \(\widehat D_{\mathrm{in}}(Z_j)\), \(\widehat D_{\mathrm{out}}(Z_j)\) are the sum of the bootstrap-support values coming into and going out from feature $j$, respectively. Similarly for CopDAG:
\begin{equation}
\label{indegree_and_outdegree_2}
\widehat D_C
=
\left\{
\left(
\widehat D_{\mathrm{in}}(C_j),
\widehat D_{\mathrm{out}}(C_j)
\right):
j=1,\ldots,p
\right\},
\end{equation}

where \(\widehat D_{\mathrm{in}}(C_j)\), \(\widehat D_{\mathrm{out}}(C_j)\) are the sum of the bootstrap-support values coming into and going out from feature $j$, respectively.

Then, for each features $j=1,...,p$, we obtain the connectivity score ($S(j) \in\{S_{Z}(j),S_{C}(j),S_{R}(j)\}$) from all three methods as follows:
\begin{equation}
\label{score_Xj_1}
\mathrm{S}_{Z}(j) = \widehat{D}_{\mathrm{in}}(Z_j) + \widehat{D}_{\mathrm{out}}(Z_j),
\end{equation}
\begin{equation}
\label{score_Xj_2}
\mathrm{S}_{C}(j) = \widehat{D}_{\mathrm{in}}(C_j) + \widehat{D}_{\mathrm{out}}(C_j),
\end{equation}
\begin{equation}
\label{score_Xj_3}
\mathrm{S}_{R}(j) = \sum_{k:(j,k)\in \hat{E'}_{R}} |\hat\tau_{jk}|.
\end{equation}

Then, the resulting scores are normalized separately to have unit mean, yielding the corresponding feature weights, $w_{Z,j}, w_{C,j}$ and $w_{R,j}$, as

\begin{equation*}
w_{a,j}
=
\frac{S_a(j)}
{\frac{1}{p}\sum_{k=1}^{p} S_a(k)},
\qquad
a\in\{Z,C,R\}, \quad j=1,\ldots,p.
\label{eq:wj=normalizas_j}
\end{equation*}

The normalized weights were then incorporated into $Z$ to obtain the corresponding standardized weighted feature representations, defined as

\begin{equation}
Z_{a,j}^{*}
=
\sqrt{w_{a,j}}\,Z_j,
\qquad
a\in\{Z,C,R\}, \quad j=1,\ldots,p.
\label{eq:weighted_standardized_features}
\end{equation}

Thus, $Z_{Z,j}^{*}$, $Z_{C,j}^{*}$, and $Z_{R,j}^{*}$ correspond to the raw-DAG, CopDAG, and R-vine weighted representations respectively. This transformation allows features with larger connectivity scores to contribute more strongly while retaining the standardized scale of the original predictor features. The square root transformation ensures the contribution of each feature to the squared Euclidean distance is proportional to its corresponding weight $w_{a,j}$, rather than $w_{a,j}^2$.

The proposed framework is summarized in Algorithm~\ref{alg:dependence_guided_clustering}.

\algrenewcommand\algorithmicrequire{\textbf{Input:}}
\algrenewcommand\algorithmicensure{\textbf{Output:}}

\begin{algorithm}[H]
\caption{Proposed CopDAG Data Representation}
\label{alg:dependence_guided_clustering}

\begin{algorithmic}[1]

\Require Data matrix, $\mathbf{X}=(X_1,\ldots,X_p) \in \Re^{N \times p}$, $\gamma$: stability threshold 
\Ensure copDAG transformed data representation $\mathbf{Z^*}$, Eq. \ref{eq:weighted_standardized_features}
\State Standardize: $\mathbf{X} \rightarrow \mathbf{Z}$
\Statex
\State \textbf{For} $b=1,\ldots,B$ do
\Statex \hspace{\algorithmicindent} Draw $n$ samples with replacement $\mathbf{Z}^{(b)}$ from $\mathbf{Z}$.
\Statex \hspace{\algorithmicindent} Obtain Pseudo-observations $\mathbf{Z}^{(b)}  \rightarrow \mathbf{P}^{(b)}$, for R-vine Eq. \ref{equation_pseudo_transform}
\Statex \hspace{\algorithmicindent} Obtain Gaussian copula scores $\mathbf{Z}^{(b)}$ $\rightarrow \mathbf{C}^{(b)}$ Eq. \ref{equation_gaussian_transform}
\Statex \hspace{\algorithmicindent} raw-DAG modeling
$\mathbf{Z}^{(b)} \rightarrow \widehat{\mathcal{G}}^{(b)}_{\mathrm{Z,\ell}}
    =
    \left(V,\widehat{E}_{\ell,\mathrm{Z}}^{(b)}\right)$,  Eq. \ref{eq:bootstrap_dags_raw}
\Statex \hspace{\algorithmicindent} Gaussian-copula DAG:
    $\mathbf{C}^{(b)} \rightarrow \widehat{\mathcal{G}}_{\mathrm{C, \ell}}^{(b)}
    =
    \left(V,\widehat{E}_{\ell,\mathrm{C}}^{(b)}\right)$, Eq. \ref{eq:bootstrap_dags_gc}
    
\Statex \hspace{\algorithmicindent} Obtain ensemble of $L$ DAGs, $\widehat{\mathcal{G}}^{(b)}_{\mathrm{Z}}$ and $\widehat{\mathcal{G}}^{(b)}_{\mathrm{C}}$~\ref{eq:bootstrap_consensus_dags_with_majority_voting} 
    \Statex \hspace{\algorithmicindent} Apply R-vine copula to $\mathbf{P}^{(b)}$ $\widehat{\mathcal{T}}^{(b)}_{R}
    =
    \left(V,\widehat{E'}_{\mathrm{R}}^{(b)}\right)$, Eq. \ref{rvine}
\Statex \textbf{End For} 
\State \text{Aggregate across the $B$ bootstrap samples,} $\widehat{\mathcal{G}}_{\mathrm{Z}}$ \ref{final_consensus_graphs_all_ensemble_1}, $\widehat{\mathcal{G}}_{\mathrm{C}}$ \ref{final_consensus_graphs_all_ensemble_2} and $\widehat{\mathcal{T}}_{R}$ \ref{final_consensus_graphs_all_ensemble_3}

\Statex
\State Obtain connectivity/dependence information:

\Statex \hspace{\algorithmicindent}
$\widehat{\mathcal{G}}_{\mathrm{Z}}
\rightarrow
\left\{
\widehat{D}_{\mathrm{in}}(Z_j),
\widehat{D}_{\mathrm{out}}(Z_j)
\right\}$,
Eq.~\ref{indegree_and_outdegree_1}

\Statex \hspace{\algorithmicindent}
$\widehat{\mathcal{G}}_{\mathrm{C}}
\rightarrow
\left\{
\widehat{D}_{\mathrm{in}}(C_j),
\widehat{D}_{\mathrm{out}}(C_j)
\right\}$,
Eq.~\ref{indegree_and_outdegree_2}

\Statex \hspace{\algorithmicindent}
$\widehat{\mathcal{T}}_{R}
\rightarrow
\left|\widehat{\tau}_{jk}\right|$,
$(j,k)\in\widehat{E}'_{R}$,
Eq.~\ref{score_Xj_3}

\Statex
\State \textbf{For} $j=1,\ldots,p$ do
\Statex \hspace{\algorithmicindent} $S(j) \in \{S_Z(j), S_C(j) , S_R(j)\} \leftarrow \widehat{D}(Z_j),\widehat{D}(C_j), \hat{\tau}_{jk}$ Eq. \ref{score_Xj_1} - \ref{score_Xj_3}
\Statex \hspace{\algorithmicindent}
For $a\in\{Z,C,R\}$, normalize weight:
$
w_{a,j}
=
\displaystyle
\frac{S_a(j)}
{\frac{1}{p}\sum_{k=1}^{p}S_a(k)}
$

\Statex \hspace{\algorithmicindent}
\hspace{\algorithmicindent} \hspace{\algorithmicindent} \hspace{\algorithmicindent} 
\hspace{\algorithmicindent} ~~~weighted feature:
$
Z_{a,j}^{*}
=
\sqrt{w_{a,j}}\,Z_j
$
\Statex \textbf{End For}
\Statex \textbf{Return} $\mathbf{Z^*}$.
\end{algorithmic}
\end{algorithm}

\subsection{Clustering Methods and Evaluation Metrics} 

Eq. \ref{eq:weighted_standardized_features} will be subject to clustering under the hypothesis that the corresponding data representation is cluster-friendly because it embeds necessary conditional and structural feature information into data.
We considered three standard clustering approaches for this study: K-means \citep{macqueen1967kmeans}, hierarchical clustering using Ward's \(D^2\)  linkage (HWD$^2$) \citep{ward1963hierarchical}, and Gaussian Mixture Models (GMM) \citep{mclachlan2000mixture} clustering.
These approaches represent three distinct clustering paradigms: distance-based, hierarchical agglomerative, and model-based clustering, respectively. Evaluating the proposed framework across these paradigms allows more consistent and comprehensive performance assessment rather than a single clustering approach. We evaluated performance using three criteria: clustering accuracy (ACC), adjusted Rand index (ARI), and normalized mutual information (NMI). 

 \textbf{ACC} finds the best matching between the predicted and the true labels using the Hungarian algorithm \citep{kuhn1955}. Let $Y_{true}^i$, ${Y}_{pred}^i$, and $m(\cdot)$ denote the true label of observation $i$, the predicted label, and the optimal label mapping, respectively. ACC is defined as
\begin{equation}
\text{ACC} = \max_m \frac{\sum_{i=1}^{N} 1(Y_{true}^i == m({Y}_{pred}^i))}{N} ,
\end{equation}
where $N$ is the number of observations. A higher ACC indicates stronger mapping between the predicted clusters and the true class labels.

 \textbf{ARI} measures the agreement between the true and predicted clusters while adjusting for agreement expected by chance \citep{hubert1985comparing}. 
ARI is defined as
\begin{equation}
\text{ARI} = \frac{RI - E(RI)}{\max(RI) - E(RI)},   \quad \mbox{with}~~~~~~~ RI = \frac{TP + TN}{\binom{N}{2}},
\end{equation} 
where $TP$ and $TN$ denote the number of true positive and true negative pairs, respectively, and $\binom{N}{2}$ is the total number of observation pairs. ARI takes a maximum value of 1 for perfect agreement, while negative values indicate agreement worse than expected by chance.

 \textbf{NMI} measures the shared information between the true class labels and predicted cluster labels from an information-theoretic perspective \citep{strehl2002} and is invariant to permutations of cluster labels \citep{strehl2002, samad2026miningelectronichealthrecords}. NMI is defined as
\begin{equation}
\text{NMI}(G,P)=2*\frac{I(G,P)},
{H(G)H(P)}
\end{equation} 

where $I(G,P)$ denotes the mutual information between the ground-truth labels $G$ and predicted cluster labels $P$, and $H(G)$ and $H(P)$ denote their respective entropies. NMI is typically scaled between 0 and 1, with 1 indicating perfect agreement between the predicted clusters and the true labels.

\subsection{Experiment: Application on Biomedical Datasets}
\label{sec:experiments}

We evaluate the performance of the proposed framework using 16 publicly available biomedical datasets. This application assesses the clustering performance across different disease domains, sample sizes, feature dimensions, and class structures. Table~\ref{tab:benchmark_datasets} summarizes the datasets used in this application. 

\begin{table}[htbp]
\centering
\caption{Summary of benchmark biomedical datasets used in the application study.  $n$ is the sample size and $p$ is the number of features.}
\label{tab:benchmark_datasets}
\normalsize
\setlength{\tabcolsep}{3pt}
\resizebox{\textwidth}{!}{
\begin{tabular}{lcclp{4cm}}
\hline
Dataset & $n$ & $p$ & Response features & Source \\
\hline
Breast Cancer Wisconsin & 699 & 9 & Cancer status  & Wolberg et al.~\citep{mangasarian1990cancer} \\
Pima Indians Diabetes & 393 & 8 & Diabetes status & Smith et al.~\citep{smith1988adap}\\
Heart Failure & 299 & 12 & Survival status & Chicco and Jurman~\citep{chicco2020heartfailure} \\
Vertebral & 310 & 6 & Orthopedic diagnosis  & UCI ML Repository \citep{barreto2005vertebral} \\
Parkinsons & 195 & 22 & Parkinson's status  & Little et al.~\citep{little2009parkinsons}\\
Heart Disease (Cleveland) & 303 & 13 & Heart disease status & OpenML Dataset\citep{openml2022heartdiseasepatients} \\
Heart Disease (Long Beach) & 200 & 13 & Heart disease status  & Janosi et al. \citep{janosi1989heart} \\
Bone Marrow Transplant & 187 & 32 & Survival status & Sikora et al.~\citep{sikora2020bonemarrow}\\
Sachs Protein-signaling & 11672 & 10 & PKA status  & Sachs et al.~\citep{sachs2005protein} \\
Breast Cancer Coimbra & 116 & 9 & Breast cancer status & Patrício et al.\citep{patricio2018breastcoimbra}  \\
Hepatitis C Virus (HCV) & 615 & 12 & Hepatitis status  & Lichtinghagen et al.~\citep{hcv_data_571}\\
Breast Tissue & 106 & 9 & Breast tissue diagnosis  & Silva et al.~\citep{uciBreastTissue} \\
Staglog Heart & 270 & 13 & Heart disease status & Statlog Project~\citep{statlog_heart_145}\\
Mammographic Mass & 961 & 5 & Breast mass severity  & Elter et al.~\citep{mammographic_mass_161} \\
Indian Liver Patient & 583 & 10 & Liver disease status  & Ramana et al.~\citep{ilpd_indian_liver_patient_dataset_225} \\
Haberman's Survival & 306 & 3 & Survival status &  Haberman~\citep{habermans_survival_43} \\
\hline
\end{tabular}
}
\end{table}

We addressed missing data by imputing the mean for continuous variables and the mode for binary variables; we removed observations with missing response variables. Supplementary material includes further information on preprocessing steps and feature exclusion. For this application, we chose the stability threshold $\gamma$ from 0.3, 0.4, and 0.6 based on the observed support levels, since bootstrap edge-support distributions differed across datasets. When support values were high, we applied higher thresholds, whereas we used lower thresholds when needed to prevent an empty dependence graph. However, the clustering results changed only marginally across these threshold settings. Consequently, this dataset-specific tuning should not be considered a limitation.

\section {Results}
\label{sec:results}
This section presents the clustering performance of the proposed CopDAG framework and two comparative feature representations—raw-DAG, R-vine, and the original unweighted raw data—across 16 biomedical datasets using various clustering algorithms.

\subsection {Best method for individual datasets}
The best-performing clustering result for each dataset is summarized in Table~\ref{tab:best_result_each_dataset}. Overall, dependence- (copula) and directional-guided  (DAG) data representations delivered the best clustering results in 11 of the 16 biomedical datasets. In nine datasets, these representations were uniquely top-performing, whereas in two datasets they tied with the raw data for best performance. Raw data alone performed best in the remaining five datasets. Among the dependence-guided representations, R-vine was chosen most often and achieved the best performance on four datasets, whereas CopDAG and raw-DAG each delivered the top result on three datasets. These results suggest that integrating feature-dependence information often yielded the most effective representation for retrieving the latent class structure across the biomedical datasets. Full clustering performance results for all 16 datasets are reported in Supplementary Tables~\ref{tab:supp_perf_all_datasets} and ~\ref{tab:supp_perf_all_datasets_2}.

\begin{table}[t]
\centering
\caption{Best clustering result within each biomedical dataset, grouped by ACC and class ratio. $p$ is the number of features, $N$ indicates approximate normality based on Shapiro--Wilk test ($p < 0.01$).}
\label{tab:best_result_each_dataset}
\normalsize
\setlength{\tabcolsep}{3pt}
\resizebox{\textwidth}{!}{
\begin{tabular}{ll l cccccccc}
\toprule
\textbf{Dataset} & \shortstack{\textbf{Best}\\\textbf{Method}} & \textbf{Clustering} & \textbf{ACC} & \textbf{ARI} & \textbf{NMI} & \shortstack{\textbf{Cont.}\\\textbf{$p$}}  & \textbf{$N$} & \textbf{\%$N$} &  \shortstack{\textbf{Class}\\\textbf{ratio}}  & \textbf{Imbalance}\\
\midrule
Breast Cancer Wisconsin 
& R-vine & K-means 
& 0.972 & 0.891 & 0.808 & 9 & 0 & 0\% & 1:1.9 & Balanced\\
Statlog Heart
& raw data & K-means 
& 0.837 & 0.452 & 0.363 & 5 & 1 & 20\% & 1:1.3 & Balanced\\
HD (Cleveland) 
& raw data & K-means 
& 0.835 & 0.447 & 0.367 & 5 & 0 & 0\% & 1:1.2 & Balanced\\
Mammographic Mass
& CopDAG & K-means 
& 0.816 & 0.398 & 0.333 & 5 & 0 & 0\% & 1:1.2 & Balanced\\
Pima Indians Diabetes 
& raw data & K-means 
& 0.746 & 0.235 & 0.158 & 8 & 0 & 0\% & 1:2.0 & Balanced\\
Breast Tissue 
& raw data & GMM 
& 0.660 & 0.414 & 0.580 & 9 & 0 & 0\% & 1:1.6 & Balanced\\
Breast Cancer Coimbra &
\makecell{raw data, raw-DAG} &
GMM &
0.638 & 0.068 & 0.112 & 9 & 0 & 0\% & 1:1.2 & Balanced \\
Bone Marrow Transplant
& raw data & K-means 
& 0.578 & 0.019 & 0.019 & 6 & 0 & 0\% & 1:1.2 & Balanced\\
Sachs Protein-signaling &
\makecell{raw data, raw-DAG} &
GMM &
0.547 & 0.022 & 0.026 & 10 & 0 & 0\% & 1:1.0 & Balanced \\
Vertebral 
& raw-DAG & GMM 
& 0.787 & 0.328 & 0.308 & 6 & 1 & 16.7\% & 1:2.1 & Mild\\
Heart Failure 
& CopDAG & GMM 
& 0.712 & 0.177 & 0.133 & 7 & 0 & 0\% & 1:2.1 & Mild\\
Haberman’s Survival
& CopDAG & GMM 
& 0.696 & 0.135 & 0.072 & 3 & 1 & 33.3\% & 1:2.8 & Mild\\
HD (Long Beach)  & R-vine & HWD$^2$  & 0.675 & 0.068 & 0.019 & 5 & 0 & 0\% & 1:2.9 & Mild\\
Indian Liver Patient
& R-vine & K-means 
& 0.617 & 0.048 & 0.107 & 9 & 0 & 0\% & 1:2.5 & Mild\\
Parkinsons 
& R-vine & GMM 
& 0.759 & 0.180 & 0.073 & 22 & 4 & 18.2\% & 1:3.1 & Moderate\\
HCV
& raw-DAG & GMM 
& 0.569 & 0.184 & 0.230 & 11 & 0 & 0\% & 1:25.7 & Very high\\
\bottomrule

\vspace{0.01cm}

\end{tabular}
}
\small\textit{Note: Multiple methods listed for a dataset indicate tied best performance.}
\end{table}

\begin{table}[htbp]
\centering
\normalsize
\setlength{\tabcolsep}{3pt}
\renewcommand{\arraystretch}{1.05}

\caption{Clustering performance across the biomedical datasets. Scores are normalized across the methods and then averaged across the datasets for comparison.
Lower rank indicates better performance.}
\label{tab:normalized_performance_avg_rank}

\resizebox{\textwidth}{!}{
\begin{tabular}{llccccccc}
\hline
& & \multicolumn{2}{c}{ACC} 
& \multicolumn{2}{c}{ARI} 
& \multicolumn{2}{c}{NMI} 
& \\
\cline{3-4}\cline{5-6}\cline{7-8}
Method & Clustering 
& Score & Rank 
& Score & Rank 
& Score & Rank 
& Avg. Rank \\
\hline

raw-DAG & GMM
& 0.628 $\pm$ 0.352 & 2
& 0.606 $\pm$ 0.320 & 3
& \textbf{0.588 $\pm$ 0.356} & \textbf{1}
& \textbf{2.0} \\

CopDAG & K-means
& \textbf{0.637 $\pm$ 0.284} & \textbf{1}
& \textbf{0.618 $\pm$ 0.315} & \textbf{1}
& 0.538 $\pm$ 0.325 & 6
& 2.7 \\

raw data & GMM
& 0.604 $\pm$ 0.403 & 4
& 0.613 $\pm$ 0.356 & 2
& 0.583 $\pm$ 0.379 & 2
& 2.7 \\

raw data & K-means
& 0.611 $\pm$ 0.405 & 3
& 0.477 $\pm$ 0.439 & 5
& 0.579 $\pm$ 0.348 & 3
& 3.7 \\

R-vine & K-means
& 0.585 $\pm$ 0.339 & 5
& 0.564 $\pm$ 0.344 & 4
& 0.549 $\pm$ 0.373 & 5
& 4.7 \\

R-vine & HWD$^2$
& 0.481 $\pm$ 0.337 & 9
& 0.457 $\pm$ 0.352 & 6
& 0.559 $\pm$ 0.349 & 4
& 6.3 \\

CopDAG & GMM
& 0.492 $\pm$ 0.357 & 7
& 0.445 $\pm$ 0.348 & 7
& 0.474 $\pm$ 0.347 & 8
& 7.3 \\

raw-DAG & HWD$^2$
& 0.530 $\pm$ 0.348 & 6
& 0.431 $\pm$ 0.365 & 9
& 0.380 $\pm$ 0.359 & 10
& 8.3 \\

CopDAG & HWD$^2$
& 0.452 $\pm$ 0.339 & 10
& 0.431 $\pm$ 0.346 & 8
& 0.510 $\pm$ 0.294 & 7
& 8.3 \\

R-vine & GMM
& 0.439 $\pm$ 0.322 & 12
& 0.380 $\pm$ 0.322 & 10
& 0.431 $\pm$ 0.317 & 9
& 10.3 \\

raw data & HWD$^2$
& 0.490 $\pm$ 0.376 & 8
& 0.365 $\pm$ 0.344 & 12
& 0.357 $\pm$ 0.345 & 12
& 10.7 \\

raw-DAG & K-means
& 0.439 $\pm$ 0.288 & 11
& 0.365 $\pm$ 0.337 & 11
& 0.376 $\pm$ 0.294 & 11
& 11.0 \\

\hline
\end{tabular}
}
\end{table}

\subsection{Overall clustering performance}
Table~\ref{tab:normalized_performance_avg_rank} presents the normalized clustering performance results for the biomedical datasets. We computed average ranks independently for ACC, ARI, and NMI, then combined them by averaging across these three measures. Among all data representations and clustering combinations, raw-DAG achieved the best overall average rank when used with GMM clustering. raw-DAG produced the highest mean NMI scores. However, CopDAG combined with k-means clustering attained the highest average ACC and ARI scores. However, it achieves the second-best overall rank because it performs worse on the NMI metric. Within each clustering method, DAG-based data representations achieved the highest average normalized ACC scores. CopDAG produced the highest ACC under K-means, while raw-DAG produced the highest ACC with HWD$^2$ and GMM clustering methods. Clustering the raw data with GMM and k-means places them third and fourth overall, although k-means ranks fifth when considering ARI scores. GMM secures the top two ranks for NMI scores, suggesting that such clustering may help preserve the mutual information of the clusters. In general, the effectiveness of the data representations depends on the selection of clustering methods and performance metrics.

\begin{figure*}[htbp]
\centering
\captionsetup{skip=4pt}

\setlength{\tabcolsep}{2pt}
\renewcommand{\arraystretch}{0.90}

\begin{tabular}{
    >{\centering\arraybackslash}m{0.17\textwidth}
    cccccc
}

\textbf{Dataset}
&
\parbox[c]{0.1\textwidth}{\centering\small\textbf{raw data}}
&
\parbox[c]{0.15\textwidth}{\centering\small\textbf{raw-DAG}}
&
\parbox[c]{0.15\textwidth}{\centering\small\textbf{CopDAG}}
&
\parbox[c]{0.15\textwidth}{\centering\small\textbf{R-vine}} &
\parbox[c]{0.1\textwidth}{\small\textbf{Best}} &
\parbox[c]{0.1\textwidth}{\small\textbf{$2^{nd}$ Best}}
\\[7pt]
\raisebox{1cm}[0pt][0pt]{%
\parbox{0.17\textwidth}{\centering\small
\textbf{Breast Cancer\\Wisconsin}}}
&
\multicolumn{4}{l}{
\includegraphics[width=0.60\textwidth]
{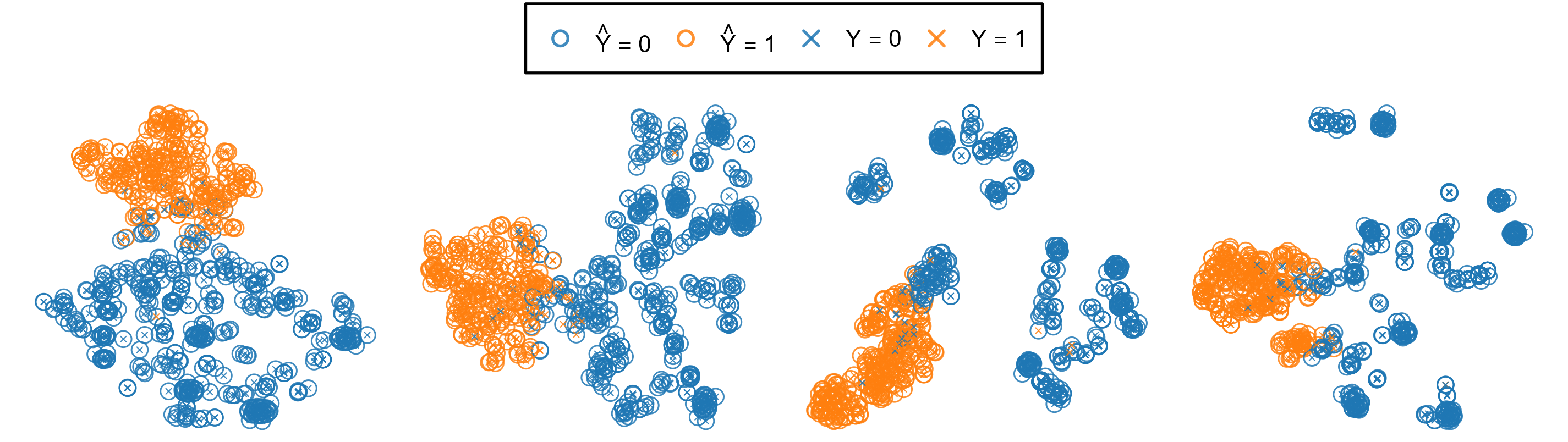}} & \raisebox{1cm}[0pt][0pt]{%
\parbox{0.1\textwidth}{\small
\textbf{R-vine}}} & \raisebox{1cm}[0pt][0pt]{%
\parbox{0.15\textwidth}{\small
\textbf{CopDAG}}} 
\\[-4pt]
\midrule
\raisebox{1cm}[0pt][0pt]{%
\parbox{0.17\textwidth}{\centering\small
\textbf{Pima Indians\\Diabetes}}}
&
\multicolumn{4}{l}{
\includegraphics[width=0.60\textwidth]
{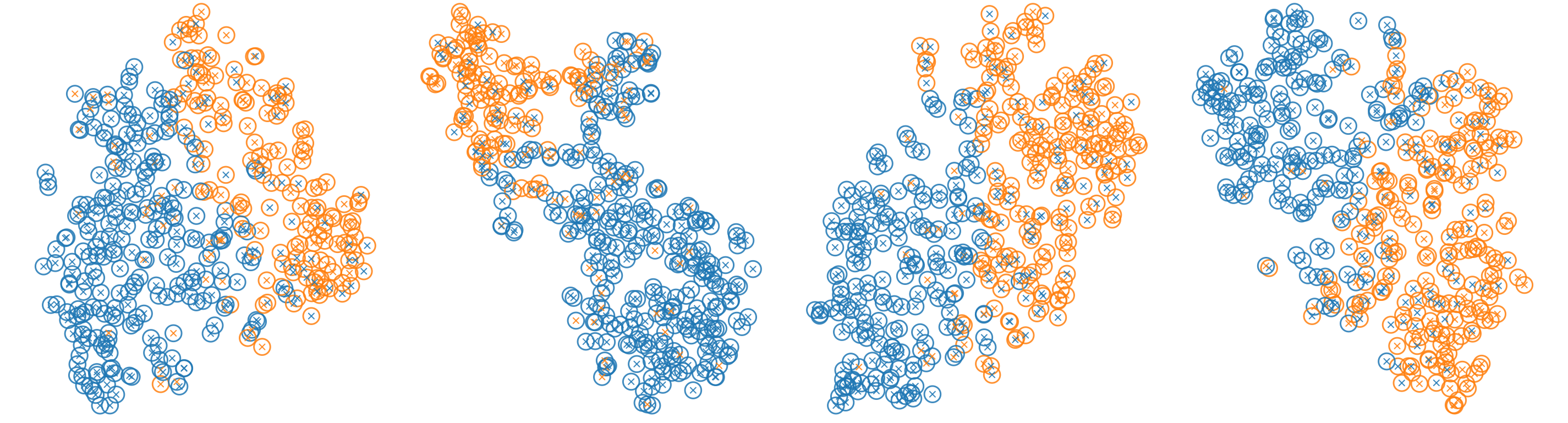}}& \raisebox{1cm}[0pt][0pt]{%
\parbox{0.1\textwidth}{\small
\textbf{raw data}}} & \raisebox{1cm}[0pt][0pt]{%
\parbox{0.1\textwidth}{\small
\textbf{raw-DAG}}} 
\\[-4pt]
\midrule
\raisebox{1cm}[0pt][0pt]{%
\parbox{0.17\textwidth}{\centering\small
\textbf{Heart Disease\\(Cleveland)}}}
&
\multicolumn{4}{c}{
\includegraphics[width=0.60\textwidth]
{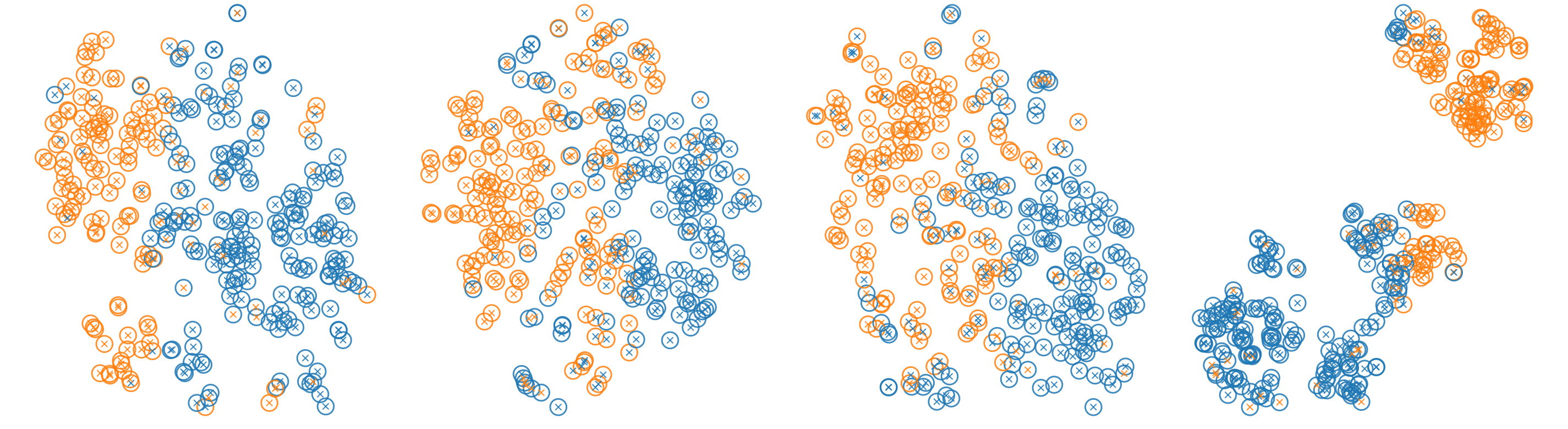}}
& \raisebox{1cm}[0pt][0pt]{%
\parbox{0.1\textwidth}{\small
\textbf{raw data}}} & \raisebox{1cm}[0pt][0pt]{%
\parbox{0.1\textwidth}{\small
\textbf{R-Vine}}} 
\end{tabular}

\caption{t-SNE visualizations using K-means clustering. Open circles indicate $\hat{Y}$ and crosses indicate true labels $Y$.}

\label{fig:tsne_kmeans_3_datasets}
\end{figure*}

\subsection{Effects of clustering methods}
Clustering approaches have influenced how effectively data are represented, whether using copula-based representations, DAG-based representations, or the raw data itself. 
GMM consistently placed among the top-performing approaches, while K-means varied more widely—ranking second overall with CopDAG but dropping to the lowest position with raw-DAG (avg. rank = 11). In contrast, HWD$^2$ tended to underperform, with every combination falling in the bottom half of the rankings. Large standard deviations in clustering performance further indicate substantial variability between datasets, supporting the use of average rank as a more reliable overall summary than mean performance alone. Figure \ref{fig:tsne_kmeans_3_datasets} compares the t-SNE visualizations for two datasets in which K-means achieved the best performance, with three additional datasets provided in Supplementary \ref{fig:tsne_kmeans_additional3_datasets}. The observed differences among the four representations suggest that the proposed copula and DAG-based transformations alter the distribution of observations in the embedded space. We have provided further t-SNE visualizations in the Supplementary material.
Given the substantial variation in performance across datasets, the change in min–max normalized $\Delta\mathrm{ACC}=\mathrm{ACC}_{\mathrm{weighted}}-\mathrm{ACC}_{\mathrm{raw}}$ was assessed using the Wilcoxon signed-rank test with Holm-adjusted $p$-values (Table~\ref{tab:acc_improvement_raw_baseline}).

\begin{table*}[htbp]
\centering
\caption{$\Delta\mathrm{ACC}$ relative to the corresponding raw-data. The improved, tied, and worsened columns report the number of datasets with positive, zero, and negative changes, respectively. }
\label{tab:acc_improvement_raw_baseline}

\normalsize
\setlength{\tabcolsep}{3pt}
\resizebox{\textwidth}{!}{
\setlength{\tabcolsep}{4pt}
\renewcommand{\arraystretch}{1.1}
\begin{tabular}{llccccccc}
\toprule
\textbf{Method} &
\textbf{Clustering} &
\makecell{\textbf{Mean}\\$\boldsymbol{\Delta}$\textbf{ACC}} &
\makecell{\textbf{Median}\\$\boldsymbol{\Delta}$\textbf{ACC}} &
\textbf{Impr.} &
\textbf{Tied} &
\textbf{Wors.} &
\makecell{\textbf{Impr.}\\\textbf{$\%$}} &
\makecell{\textbf{Holm-adj}\\$\boldsymbol{p}$\textbf{-value}} \\
\midrule

R-vine
& K-means
& -0.0268
& 0.0678
& 11
& 0
& 5
& 68.8\%
& 1.0000 \\

CopDAG
& K-means
& 0.0257
& 0.0949
& 10
& 0
& 6
& 62.5\%
& 1.0000 \\

raw-DAG
& K-means
& -0.1721
& -0.0583
& 3
& 3
& 10
& 18.8\%
& 0.3843 \\
\midrule
R-vine
& HWD$^2$
& -0.0085
& 0.0891
& 9
& 0
& 7
& 56.3\%
& 1.0000 \\

CopDAG
& HWD$^2$
& -0.0380
& 0.0398
& 8
& 0
& 8
& 50.0\%
& 1.0000 \\

raw-DAG
& HWD$^2$
& 0.0405
& 0.0045
& 8
& 1
& 7
& 50.0\%
& 1.0000 \\
\midrule 
CopDAG
& GMM
& -0.1528
& -0.0615
& 6
& 1
& 8
& 40.0\%
& 1.0000 \\

R-vine
& GMM
& -0.2054
& -0.0615
& 5
& 1
& 9
& 33.3\%
& 1.0000 \\

raw-DAG
& GMM
& 0.0243
& 0.0000
& 5
& 8
& 3
& 31.3\%
& 1.0000 \\
\bottomrule
\end{tabular}
}
\end{table*}

The largest gains in improvement frequency relative to clustering on raw data were seen with K-means. Compared with the raw-data baseline, R-vine increased ACC on 11 of 16 datasets (68.8$\%$), whereas CopDAG increased ACC on 10 datasets (62.5$\%$). The positive median $\Delta\mathrm{ACC}$ suggests that improvements were achieved for most datasets, but these were counterbalanced by larger drops in a smaller subset. In summary, both CopDAG and R-vine were most reliably advantageous when paired with K-means. 
However, the gains were not consistent across datasets, and after applying the Holm correction, none of the comparisons remained statistically significant. This indicates that any advantage of dependence-guided weighting may be contingent on properties of the underlying data rather than reliably observed across all datasets.

\begin{figure*}[!t]
\centering
\setlength{\tabcolsep}{3pt}
\renewcommand{\arraystretch}{0.95}
\begin{tabular}{ccc}

\multicolumn{3}{c}{
\textbf{Class 0: Non-diabetic (\(n=263\))}
}
\\[1mm]

\includegraphics[width=0.30\textwidth]
{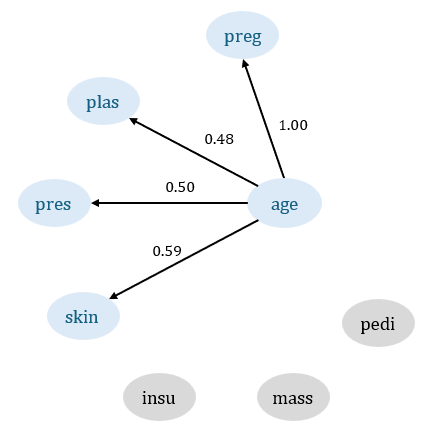}
&
\includegraphics[width=0.30\textwidth]
{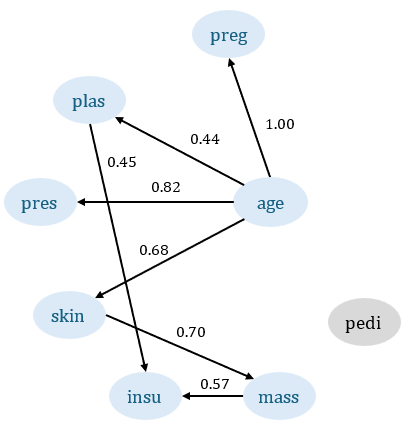}
&
\includegraphics[width=0.30\textwidth]
{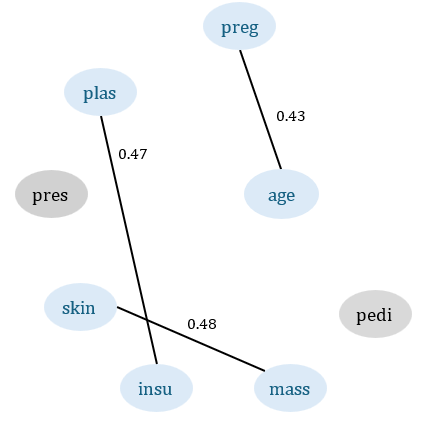}
\\[2mm]
\multicolumn{3}{c}{
\textbf{Class 1: Diabetic (\(n=130\))}
}
\\[1mm]

\includegraphics[width=0.30\textwidth]
{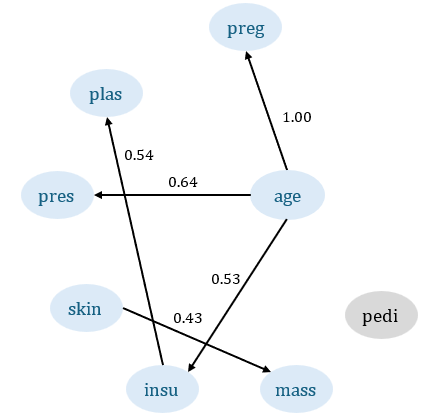}
&
\includegraphics[width=0.30\textwidth]
{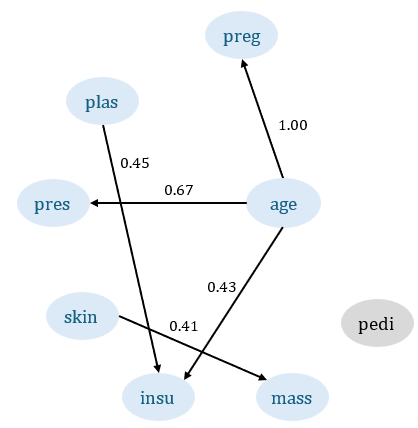}
&
\includegraphics[width=0.30\textwidth]
{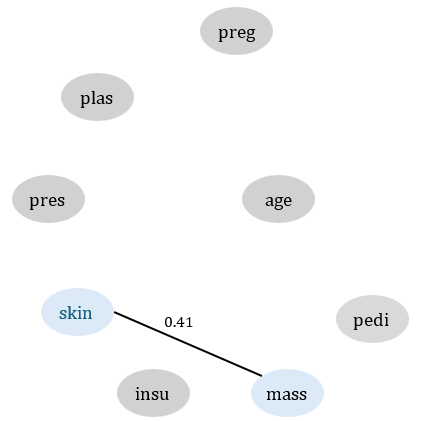}
\\[2mm]

\small (a) raw-DAG
&
\small (b) CopDAG
&
\small (c) R-vine

\end{tabular}

\caption{Class-specific dependence structures for the Pima Indians Diabetes dataset. 
}
\label{fig:diabetes_class_specific_dependence_graphs}

\end{figure*}

\subsection{Feature dependence structures}

To improve the explainability of the proposed data representation approaches, we can use DAG and R-vine structures to visualize dependencies among biomedical features. The class-specific feature dependence graphs for the Pima Indians Diabetes dataset provide some information on this issue (Figure~\ref{fig:diabetes_class_specific_dependence_graphs}). The causal DAG structure on the raw data clearly shows that, for non-diabetic patients, most clinical features depend only on age. However, in patients with diabetes, the direct structure of feature dependence changes and becomes more complex, which may help explain why the transformed data representation distinguishes patient classes by feature dependence.  Similarly, copDAG shows distinct causal and directed graph structures for diabetic and non-diabetic patients. The R-vine copula show different structures for diabetic and non-diabetic patients. In addition to the distinct feature-dependence structures, the associated dependence weights shed further light on disease- and cohort-specific characteristics. These differences suggest that class separation may be influenced not only by marginal feature values but also by class-specific dependence patterns among features.

\section{Discussion}

This paper introduces a novel data representation approach that combines multivariate copula modeling with conditional dependencies modeled using causal directed graphs. The findings of the paper can be summarized as follows. First, the DAG-based causal dependence model achieved the top rank, and CopDAG achieved the best data representations in terms of ACC and ARI scores. Specifically, CopDAG with K-means achieved the highest overall average ACC and ARI among all feature-representation and clustering combinations. Second, the effectiveness of data representation methods depends on the dataset and the clustering technique. In other words, to outperform baseline clustering of raw data, inductive alignment among data representations, the dataset, and clustering methods is needed. Third, the largest gains in clustering our proposed data representations against raw data is observed when K-means is used as the clustering method. Fourth, the directed graph structures of the causal DAG models and the vine structures of the R-vine also demonstrate distinct relationships between features for different cluster or patient cohorts. Fifth, although clustering performance improves with causal DAG-based data representations over raw data, incorporating feature dependence via a Gaussian copula transformation generally further enhances performance.

Clustering is a challenging machine learning problem because the model assumes class labels are completely unavailable. Therefore, discovering clusters related to unknown classes from complex feature dependencies is an important research problem. In this setting, the feature space may be high-dimensional, making feature relationships and dependency patterns more complex. As a result, distance-based clustering approaches, such as k-means, or distribution-based methods, such as GMM, may struggle to detect clusters directly from raw data. Therefore, a new, compact representation of the raw feature space, characterized by conditional dependencies between features, can help separate otherwise overlapping clusters in the raw data. DAG-based causal models discover directional relationships between features with corresponding weights, which we assume vary across cluster observations. This observation explains why DAG-based data-representation methods appear so strongly against clustering of raw data. 

In machine learning, no single model performs best or is the optimal choice for all data problems. Datasets exhibit diverse statistical complexities and inductive biases, so data-specific learning models are typically advised, and the most effective approach for a particular data problem is often identified through experimentation. The theoretical relationship between the clustering and data representation approach is unclear and is further complicated by the statistical properties of the input dataset. That is why data representation methods do not perform consistently across datasets and clustering methods. Although the improvement in clustering accuracy is not statistically significant, a 5\% increase in clustering accuracy is equivalent to 50 of 1000 observations, which has practical value in science.

Several contributing factors can be linked to the integrated approach based on DAGs and copulas. First, biomedical variables are rarely independent. Second, biological and clinical outcomes are rarely determined by a single cause; instead, they usually arise from interactions among multiple factors and underlying mechanisms. Third, standard modeling assumptions, estimation uncertainty, high dimensionality, and sparse data can constrain the performance of any single analytical method. Within a DAG representation, variables are represented as nodes connected according to their estimated conditional dependence relationships. These structural relationships can then guide subsequent dependence modeling and clustering. The R-vine copula component provides additional flexibility by representing complex multivariate and potentially nonlinear dependence structures without requiring all relationships to follow a single restrictive distributional form. Taken together, these elements enable simultaneous assessment of covariate dependence and the detection of informative clusters or disease-associated patterns. Finally, although the methods presented in this study are primarily motivated by biomedical applications, their utility is not restricted to the datasets considered here. Many of these techniques can be extended to other high-throughput, high-dimensional data settings, including genomic, transcriptomic, proteomic, and other omics data, where complex dependence structures, data sparsity, and limited or unreliable class labels are common analytical challenges.

\subsection{Limitations}

Despite promising results, the proposed methods have some limitations that need to be addressed in the future. Although directed-graph degree measures can be helpful, they are fairly simple and do not comprehensively capture the complex relationships among features. Therefore, more comprehensive ways to translate causal structures in DAGs into cluster-friendly data representation remain an open problem. The way to integrate copula-based feature relationships with DAG-based causal structure discovery may not be optimal, leaving room for more sophisticated model integration aligned with the clustering objective.  A deeper analysis is required to understand why dependence weighting improves performance on some datasets more than on others. Future studies will explore dataset properties—such as the strength and stability of dependence, sparsity, class-specific differences, dimensionality, and cluster overlap—to determine when each representation works best. 

The CopDAG approach requires careful, informed choices for the bootstrap sample size, majority-vote cutoff, stability threshold, DAG learning algorithms, the set of candidate copula families, and the clustering procedure. The clustering performance may be sensitive to these choices, leading to varying dependence structures and feature weights. Potential future developments of CopDAG may explore data-driven approaches for selecting stability thresholds, dependence measures that account for sign, alternative metrics for node connectivity, DAG models with mixed feature types, and combined methods that more explicitly fuse DAG structural information with R-vine dependence strength.

\section{Conclusion}

In this paper, we present a new data representation shaped by feature dependencies learned through DAG-based causal structure discovery and multivariate copula models. DAG-based feature-dependence-guided representations achieved strong recovery of the true cluster structures across different clustering settings. Clustering performance could be improved further by incorporating multivariate copula modeling of the underlying data distributions, as in CopDAG. Overall, CopDAG offers a flexible and interpretable framework for building dependence-guided representations of heterogeneous biomedical data.\\

\noindent \textbf{Declaration of Generative AI and AI-Assisted Technologies}

\noindent The authors did not use generative AI or AI-assisted technologies to develop the research methodology, models, generate data/images/charts, conduct statistical analyses, or interpret the results. AI-assisted tools were used for language editing, such as grammar and spelling corrections, and to create some illustrative elements of the graphical statistics. The authors reviewed and edited the output as needed and take full responsibility for the content of the published article.\\

\noindent \textbf{Funding}\\
The research reported in this publication received support from the US National Science Foundation (NSF) award \# 2431058.\\

\noindent \textbf{Data Availability}\\
All datasets used in this study are publicly available through the UCI Machine Learning Repository and OpenML. The corresponding sources for each dataset are provided in Table~\ref{tab:benchmark_datasets}.\\

\noindent \textbf{Conflicts of Interest}\\
The authors declare no conflicts of interest.\\

\noindent \textbf{Author Contributions}\\
All authors contributed equally to the conception, development, analysis, and preparation of this manuscript.

\bibliographystyle{elsarticle-num} 
\bibliography{ref}

\newpage
\section*{Supplementary Materials}

\vspace{-0.2cm}

\setcounter{table}{0}
\renewcommand{\thetable}{S\arabic{table}}

\setcounter{figure}{0}
\renewcommand{\thefigure}{S\arabic{figure}}
\setcounter{subsection}{0}
\renewcommand{\thesubsection}{S\arabic{subsection}}
\subsection{Clustering Performance for Biomedical Datasets}

\begin{table}[H]
\centering
\caption{Clustering performance summary for nine of the 16 biomedical datasets.}
\label{tab:supp_perf_all_datasets_2}
\tiny
\setlength{\tabcolsep}{3pt}
\renewcommand{\arraystretch}{0.85}
\begin{subtable}[t]{0.32\linewidth}
\centering
\caption{Breast Cancer}
\begin{adjustbox}{max width=\linewidth}
\begin{tabular}{llccc}
\toprule
Method & Alg. & ACC & ARI & NMI \\
\midrule
R-vine & K-means & 0.9722 & 0.8909 & 0.8078 \\
raw-DAG & HWD$^2$ & 0.9693 & 0.8798 & 0.7921 \\
CopDAG & K-means & 0.9678 & 0.8743 & 0.7889 \\
raw data & HWD$^2$ & 0.9678 & 0.8742 & 0.7839 \\
R-vine & HWD$^2$ & 0.9649 & 0.8633 & 0.7716 \\
raw data & K-means & 0.9575 & 0.8356 & 0.7335 \\
CopDAG & HWD$^2$ & 0.9561 & 0.8309 & 0.7399 \\
raw-DAG & K-means & 0.9429 & 0.7818 & 0.6747 \\
R-vine & GMM & 0.8917 & 0.6128 & 0.5935 \\
raw data & GMM & 0.8653 & 0.5331 & 0.5388 \\
raw-DAG & GMM & 0.8624 & 0.5246 & 0.5331 \\
CopDAG & GMM & 0.8082 & 0.3787 & 0.4417 \\
\bottomrule
\end{tabular}
\end{adjustbox}
\end{subtable}
\hspace{0.0001\linewidth}
\begin{subtable}[t]{0.32\linewidth}
\centering
\caption{Diabetes}
\begin{adjustbox}{max width=\linewidth}
\begin{tabular}{llccc}
\toprule
Method & Alg. & ACC & ARI & NMI \\
\midrule
raw data & K-means & 0.7455 & 0.2350 & 0.1579 \\
raw-DAG & K-means & 0.7201 & 0.1731 & 0.0925 \\
raw data & HWD$^2$ & 0.7074 & 0.1651 & 0.1033 \\
CopDAG & K-means & 0.7048 & 0.1658 & 0.1634 \\
R-vine & K-means & 0.6972 & 0.1526 & 0.1759 \\
CopDAG & HWD$^2$ & 0.6947 & 0.1411 & 0.0790 \\
R-vine & GMM & 0.6921 & 0.1455 & 0.1446 \\
CopDAG & GMM & 0.6896 & 0.1416 & 0.1423 \\
raw-DAG & GMM & 0.6896 & 0.1415 & 0.1456 \\
raw data & GMM & 0.6819 & 0.1304 & 0.1235 \\
raw-DAG & HWD$^2$ & 0.6692 & 0.0963 & 0.0433 \\
R-vine & HWD$^2$ & 0.6514 & 0.0848 & 0.1443 \\
\bottomrule
\end{tabular}
\end{adjustbox}
\end{subtable}
\hspace{0.0001\linewidth}
\begin{subtable}[t]{0.32\linewidth}
\centering
\caption{Heart Failure}
\begin{adjustbox}{max width=\linewidth}
\begin{tabular}{llccc}
\toprule
Method & Alg. & ACC & ARI & NMI \\
\midrule
CopDAG & GMM & 0.7124 & 0.1766 & 0.1326 \\
CopDAG & K-means & 0.7023 & 0.1611 & 0.1373 \\
CopDAG & HWD$^2$ & 0.6656 & 0.1026 & 0.0618 \\
raw-DAG & GMM & 0.5585 & 0.0072 & 0.0028 \\
raw data & GMM & 0.5585 & 0.0021 & 0.0005 \\
R-vine & HWD$^2$ & 0.5585 & -0.0054 & 0.0001 \\
raw-DAG & K-means & 0.5552 & -0.0019 & 0.0000 \\
R-vine & K-means & 0.5552 & -0.0019 & 0.0000 \\
R-vine & GMM & 0.5518 & -0.0109 & 0.0013 \\
raw data & K-means & 0.5485 & -0.0037 & 0.0000 \\
raw-DAG & HWD$^2$ & 0.5284 & -0.0166 & 0.0058 \\
raw data & HWD$^2$ & 0.5050 & -0.0048 & 0.0026 \\
\bottomrule
\end{tabular}
\end{adjustbox}
\end{subtable}
\vspace{0.15cm}

\begin{subtable}[t]{0.32\linewidth}
\centering
\caption{Vertebral}
\begin{adjustbox}{max width=\linewidth}
\begin{tabular}{llccc}
\toprule
Method & Alg. & ACC & ARI & NMI \\
\midrule
raw-DAG & GMM & 0.7871 & 0.3277 & 0.3080 \\
raw data & GMM & 0.7774 & 0.3058 & 0.2804 \\
raw data & HWD$^2$ & 0.7323 & 0.2122 & 0.1629 \\
R-vine & HWD$^2$ & 0.6903 & 0.1392 & 0.2057 \\
CopDAG & K-means & 0.6871 & 0.1373 & 0.1433 \\
raw-DAG & HWD$^2$ & 0.6806 & 0.1176 & 0.0615 \\
R-vine & K-means & 0.6774 & 0.1231 & 0.1306 \\
raw data & K-means & 0.6645 & 0.1019 & 0.1608 \\
raw-DAG & K-means & 0.6581 & 0.0953 & 0.1281 \\
CopDAG & GMM & 0.5742 & -0.0634 & 0.0895 \\
R-vine & GMM & 0.5613 & -0.0653 & 0.0988 \\
CopDAG & HWD$^2$ & 0.5516 & -0.0242 & 0.1303 \\
\bottomrule
\end{tabular}
\end{adjustbox}
\end{subtable}
\hspace{0.0001\linewidth}
\begin{subtable}[t]{0.32\linewidth}
\centering
\caption{Parkinsons}
\begin{adjustbox}{max width=\linewidth}
\begin{tabular}{llccc}
\toprule
Method & Alg. & ACC & ARI & NMI \\
\midrule
R-vine & GMM & 0.7590 & 0.1795 & 0.0734 \\
CopDAG & GMM & 0.7590 & 0.1732 & 0.0697 \\
CopDAG & HWD$^2$ & 0.7231 & 0.1859 & 0.1282 \\
R-vine & HWD$^2$ & 0.7179 & 0.1845 & 0.1670 \\
CopDAG & K-means & 0.7026 & 0.1602 & 0.1643 \\
R-vine & K-means & 0.6923 & 0.1440 & 0.1454 \\
raw data & HWD$^2$ & 0.6821 & 0.1267 & 0.2422 \\
raw-DAG & GMM & 0.6769 & 0.1154 & 0.0803 \\
raw-DAG & HWD$^2$ & 0.6564 & -0.0849 & 0.0666 \\
raw data & K-means & 0.6000 & -0.0978 & 0.0970 \\
raw-DAG & K-means & 0.6000 & -0.0978 & 0.0970 \\
raw data & GMM & 0.5590 & 0.0082 & 0.0295 \\
\bottomrule
\end{tabular}
\end{adjustbox}
\end{subtable}
\hspace{0.0001\linewidth}
\begin{subtable}[t]{0.32\linewidth}
\centering
\caption{Heart Disease (Cleveland)}
\begin{adjustbox}{max width=\linewidth}
\begin{tabular}{llccc}
\toprule
Method & Alg. & ACC & ARI & NMI \\
\midrule
R-vine & K-means & 0.8152 & 0.3953 & 0.3124 \\
CopDAG & K-means & 0.7888 & 0.3314 & 0.2566 \\
raw-DAG & HWD$^2$ & 0.7822 & 0.3160 & 0.2585 \\
CopDAG & GMM & 0.7657 & 0.2799 & 0.2117 \\
CopDAG & HWD$^2$ & 0.7657 & 0.2795 & 0.2638 \\
raw data & GMM & 0.7624 & 0.2729 & 0.2061 \\
raw-DAG & GMM & 0.7624 & 0.2729 & 0.2061 \\
raw-DAG & K-means & 0.7459 & 0.2393 & 0.1801 \\
raw data & HWD$^2$ & 0.7393 & 0.2264 & 0.1691 \\
R-vine & GMM & 0.7294 & 0.2077 & 0.1942 \\
R-vine & HWD$^2$ & 0.7162 & 0.1839 & 0.1460 \\
\bottomrule
\end{tabular}
\end{adjustbox}
\end{subtable}
\vspace{0.15cm}

\begin{subtable}{0.32\linewidth}
\centering
\caption{Heart Disease (Long Beach)}
\begin{adjustbox}{max width=\linewidth}
\begin{tabular}{llccc}
\toprule
Method & Alg. & ACC & ARI & NMI \\
\midrule
CopDAG & K-means & 0.6700 & 0.0631 & 0.0170 \\
CopDAG & HWD$^2$ & 0.6700 & 0.0631 & 0.0170 \\
R-vine & K-means & 0.6700 & 0.0631 & 0.0170 \\
raw data & GMM & 0.6700 & 0.0586 & 0.0146 \\
raw-DAG & GMM & 0.6700 & 0.0586 & 0.0146 \\
CopDAG & GMM & 0.6700 & 0.0586 & 0.0146 \\
R-vine & GMM & 0.6700 & 0.0586 & 0.0146 \\
raw data & K-means & 0.6650 & 0.0624 & 0.0177 \\
raw-DAG & K-means & 0.6650 & 0.0624 & 0.0177 \\
raw-DAG & HWD$^2$ & 0.6650 & 0.0538 & 0.0129 \\
raw data & HWD$^2$ & 0.6350 & 0.0363 & 0.0089 \\
\bottomrule
\end{tabular}
\end{adjustbox}
\end{subtable}
\hspace{0.0001\linewidth}
\begin{subtable}{0.32\linewidth}
\centering
\caption{Bone Marrow}
\begin{adjustbox}{max width=\linewidth}
\begin{tabular}{llccc}
\toprule
Method & Alg. & ACC & ARI & NMI \\
\midrule
raw data & K-means & 0.5775 & 0.0188 & 0.0187 \\
raw data & GMM & 0.5775 & 0.0188 & 0.0151 \\
raw data & HWD$^2$ & 0.5722 & 0.0155 & 0.0126 \\
raw-DAG & GMM & 0.5615 & 0.0098 & 0.0094 \\
raw-DAG & HWD$^2$ & 0.5615 & 0.0093 & 0.0067 \\
R-vine & HWD$^2$ & 0.5348 & -0.0020 & 0.0001 \\
raw-DAG & K-means & 0.5187 & -0.0040 & 0.0010 \\
CopDAG & K-means & 0.5187 & -0.0040 & 0.0010 \\
R-vine & K-means & 0.5187 & -0.0040 & 0.0010 \\
CopDAG & GMM & 0.5134 & -0.0046 & 0.0005 \\
R-vine & GMM & 0.5134 & -0.0046 & 0.0005 \\
CopDAG & HWD$^2$ & 0.5080 & -0.0052 & 0.0007 \\
\bottomrule
\end{tabular}
\end{adjustbox}
\end{subtable}
\hspace{0.0001\linewidth}
\begin{subtable}{0.32\linewidth}
\centering
\caption{Sachs}
\begin{adjustbox}{max width=\linewidth}
\begin{tabular}{llccc}
\toprule
Method & Alg. & ACC & ARI & NMI \\
\midrule
raw data & GMM & 0.5737 & 0.0216 & 0.0259 \\
raw-DAG & GMM & 0.5737 & 0.0216 & 0.0259 \\
raw-DAG & HWD$^2$ & 0.5666 & 0.0177 & 0.0349 \\
CopDAG & HWD$^2$ & 0.5565 & 0.0127 & 0.0309 \\
CopDAG & GMM & 0.5544 & 0.0118 & 0.0311 \\
R-vine & GMM & 0.5544 & 0.0118 & 0.0311 \\
R-vine & HWD$^2$ & 0.5525 & 0.0110 & 0.0320 \\
CopDAG & K-means & 0.5468 & 0.0087 & 0.0176 \\
raw-DAG & K-means & 0.5458 & 0.0083 & 0.0285 \\
R-vine & K-means & 0.5454 & 0.0082 & 0.0166 \\
raw data & HWD$^2$ & 0.5450 & 0.0080 & 0.0145 \\
raw data & K-means & 0.5429 & 0.0073 & 0.0276 \\
\bottomrule
\end{tabular}
\end{adjustbox}
\end{subtable}
\vspace{0.15cm}
\end{table}

\newpage 

\begin{table}[t]
\centering
\caption{Clustering performance summary for seven of the 16 biomedical datasets.}
\label{tab:supp_perf_all_datasets}
\tiny
\setlength{\tabcolsep}{3pt}
\renewcommand{\arraystretch}{0.85}
\begin{subtable}{0.32\linewidth}
\centering
\caption{Breast Cancer Coimbra}
\begin{adjustbox}{max width=\linewidth}
\begin{tabular}{llccc}
\toprule
Method & Alg. & ACC & ARI & NMI \\
\midrule
raw-DAG & GMM & 0.6379 & 0.0677 & 0.1119 \\
raw data & GMM & 0.6379 & 0.0677 & 0.1119 \\
CopDAG & K-means & 0.6207 & 0.0501 & 0.0445 \\
R-vine & K-means & 0.6121 & 0.0420 & 0.0407 \\
R-vine & HWD$^2$ & 0.6034 & 0.0345 & 0.0356 \\
raw data & K-means & 0.5603 & 0.0058 & 0.0191 \\
CopDAG & HWD$^2$ & 0.5603 & 0.0057 & 0.0061 \\
raw-DAG & K-means & 0.5431 & -0.0025 & 0.0808 \\
CopDAG & GMM & 0.5345 & -0.0042 & 0.0006 \\
R-vine & GMM & 0.5345 & -0.0042 & 0.0006 \\
raw data & HWD$^2$ & 0.5259 & -0.0060 & 0.0027 \\
raw-DAG & HWD$^2$ & 0.5086 & -0.0099 & 0.0466 \\
\bottomrule
\end{tabular}
\end{adjustbox}
\end{subtable}
\hspace{0.0001\linewidth}
\begin{subtable}{0.32\linewidth}
\centering
\caption{Hepatitis C Virus (HCV)}
\begin{adjustbox}{max width=\linewidth}
\begin{tabular}{llccc}
\toprule
Method & Alg. & ACC & ARI & NMI \\
\midrule
raw-DAG & GMM & 0.5691 & 0.1839 & 0.2303 \\
raw data & GMM & 0.5675 & 0.1908 & 0.2318 \\
raw-DAG & HWD$^2$ & 0.5642 & 0.1724 & 0.2106 \\
raw data & HWD$^2$ & 0.5593 & 0.0854 & 0.1469 \\
raw data & K-means & 0.5528 & 0.1043 & 0.1760 \\
raw-DAG & K-means & 0.4943 & 0.1411 & 0.1848 \\
CopDAG & K-means & 0.4146 & 0.0795 & 0.1496 \\
CopDAG & GMM & 0.4033 & 0.1055 & 0.1985 \\
R-vine & GMM & 0.4000 & 0.1033 & 0.2057 \\
R-vine & K-means & 0.3870 & 0.1058 & 0.2136 \\
CopDAG & HWD$^2$ & 0.3740 & 0.0495 & 0.1947 \\
R-vine & HWD$^2$ & 0.3675 & 0.0997 & 0.2186 \\
\bottomrule
\end{tabular}
\end{adjustbox}
\end{subtable}
\hspace{0.0001\linewidth}
\begin{subtable}{0.32\linewidth}
\centering
\caption{Breast Tissue}
\begin{adjustbox}{max width=\linewidth}
\begin{tabular}{llccc}
\toprule
Method & Alg. & ACC & ARI & NMI \\
\midrule
raw data & GMM & 0.6604 & 0.4143 & 0.5796 \\
raw-DAG & GMM & 0.6415 & 0.3859 & 0.5666 \\
CopDAG & K-means & 0.5660 & 0.3722 & 0.5073 \\
CopDAG & GMM & 0.5472 & 0.3419 & 0.5268 \\
R-vine & GMM & 0.5472 & 0.3419 & 0.5268 \\
raw-DAG & HWD$^2$ & 0.5472 & 0.3231 & 0.5420 \\
R-vine & HWD$^2$ & 0.5377 & 0.3238 & 0.4931 \\
raw-DAG & K-means & 0.5283 & 0.2922 & 0.5150 \\
R-vine & K-means & 0.5094 & 0.3106 & 0.4607 \\
raw data & HWD$^2$ & 0.5000 & 0.3139 & 0.5199 \\
raw data & K-means & 0.4811 & 0.2851 & 0.5234 \\
CopDAG & HWD$^2$ & 0.4717 & 0.2942 & 0.5139 \\
\bottomrule
\end{tabular}
\end{adjustbox}
\end{subtable}
\vspace{0.15cm}

\begin{subtable}{0.32\linewidth}
\centering
\caption{Statlog Heart}
\begin{adjustbox}{max width=\linewidth}
\begin{tabular}{llccc}
\toprule
Method & Alg. & ACC & ARI & NMI \\
\midrule
raw data & K-means & 0.8370 & 0.4521 & 0.3634 \\
R-vine & K-means & 0.7926 & 0.3400 & 0.2608 \\
raw data & GMM & 0.7630 & 0.2738 & 0.2051 \\
raw-DAG & GMM & 0.7630 & 0.2738 & 0.2051 \\
R-vine & HWD$^2$ & 0.7556 & 0.2584 & 0.1928 \\
CopDAG & GMM & 0.7481 & 0.2433 & 0.2242 \\
R-vine & GMM & 0.7481 & 0.2433 & 0.2242 \\
CopDAG & K-means & 0.7370 & 0.2219 & 0.1695 \\
raw data & HWD$^2$ & 0.7296 & 0.2080 & 0.1642 \\
raw-DAG & K-means & 0.7148 & 0.1816 & 0.1430 \\
CopDAG & HWD$^2$ & 0.7074 & 0.1690 & 0.1272 \\
raw-DAG & HWD$^2$ & 0.5963 & 0.0315 & 0.0688 \\
\bottomrule
\end{tabular}
\end{adjustbox}
\end{subtable}
\hspace{0.0001\linewidth}
\begin{subtable}{0.32\linewidth}
\centering
\caption{Mammographic Mass}
\begin{adjustbox}{max width=\linewidth}
\begin{tabular}{llccc}
\toprule
Method & Alg. & ACC & ARI & NMI \\
\midrule
CopDAG & K-means & 0.8158 & 0.3983 & 0.3333 \\
R-vine & K-means & 0.8137 & 0.3931 & 0.3095 \\
CopDAG & HWD$^2$ & 0.8106 & 0.3852 & 0.3314 \\
raw data & K-means & 0.7867 & 0.3280 & 0.2732 \\
R-vine & HWD$^2$ & 0.7856 & 0.3256 & 0.2900 \\
raw-DAG & HWD$^2$ & 0.7815 & 0.3162 & 0.2450 \\
raw-DAG & K-means & 0.7794 & 0.3115 & 0.2547 \\
raw data & HWD$^2$ & 0.7638 & 0.2775 & 0.2302 \\
CopDAG & GMM & 0.7294 & 0.2093 & 0.2322 \\
R-vine & GMM & 0.6753 & 0.1212 & 0.1752 \\
raw-DAG & GMM & 0.5109 & -0.0034 & 0.0041 \\
raw data & GMM & 0.5057 & -0.0037 & 0.0059 \\
\bottomrule
\end{tabular}
\end{adjustbox}
\end{subtable}
\hspace{0.0001\linewidth}
\begin{subtable}{0.32\linewidth}
\centering
\caption{Indian Liver Patient}
\begin{adjustbox}{max width=\linewidth}
\begin{tabular}{llccc}
\toprule
Method & Alg. & ACC & ARI & NMI \\
\midrule
R-vine & K-means & 0.6175 & 0.0483 & 0.1065 \\
raw data & K-means & 0.6141 & -0.0718 & 0.0637 \\
raw data & HWD$^2$ & 0.6106 & -0.0729 & 0.0659 \\
CopDAG & K-means & 0.6089 & 0.0441 & 0.0708 \\
raw-DAG & HWD$^2$ & 0.6089 & 0.0400 & 0.0187 \\
raw-DAG & K-means & 0.6003 & 0.0357 & 0.0197 \\
R-vine & HWD$^2$ & 0.5763 & 0.0061 & 0.0955 \\
CopDAG & HWD$^2$ & 0.5695 & -0.0003 & 0.0946 \\
raw data & GMM & 0.5455 & -0.0227 & 0.0979 \\
raw-DAG & GMM & 0.5455 & -0.0227 & 0.0979 \\
CopDAG & GMM & - & - & - \\
R-vine & GMM & - & - & - \\
\bottomrule
\end{tabular}
\end{adjustbox}
\end{subtable}
\vspace{0.5cm}
\newpage 
\begin{subtable}{0.32\linewidth}
\centering
\caption{Haberman's Survival}
\begin{adjustbox}{max width=\linewidth}
\begin{tabular}{llccc}
\toprule
Method & Alg. & ACC & ARI & NMI \\
\midrule
CopDAG & GMM & 0.6961 & 0.1352 & 0.0716 \\
R-vine & GMM & 0.6961 & 0.1352 & 0.0716 \\
raw data & GMM & 0.6667 & 0.1040 & 0.0704 \\
raw-DAG & GMM & 0.6667 & 0.1040 & 0.0704 \\
CopDAG & K-means & 0.6242 & 0.0561 & 0.0372 \\
R-vine & K-means & 0.6242 & 0.0561 & 0.0372 \\
CopDAG & HWD$^2$ & 0.5817 & 0.0229 & 0.0423 \\
R-vine & HWD$^2$ & 0.5817 & 0.0229 & 0.0423 \\
raw data & HWD$^2$ & 0.5556 & 0.0080 & 0.0047 \\
raw-DAG & HWD$^2$ & 0.5556 & 0.0080 & 0.0047 \\
raw data & K-means & 0.5261 & -0.0003 & 0.0010 \\
raw-DAG & K-means & 0.5261 & -0.0003 & 0.0010 \\
\bottomrule
\end{tabular}
\end{adjustbox}
\end{subtable}
\hspace{0.0001\linewidth}
\vspace{0.5cm}\\
\end{table}

\newpage
\subsection{Dataset-Specific Preprocessing}
\label{sec:supp_preprocessing}

The non-informative identifier features, including \texttt{id} and name fields, were excluded before analysis. The original Pima Indians Diabetes dataset contains 768 subjects. Observations with zero values for \texttt{skin} (triceps skin fold thickness), \texttt{insu}(two-hour serum insulin level), \texttt{pres}(diastolic blood pressure), or \texttt{mass} (body mass index) were removed because these values are not physiologically meaningful, leaving 393 observations for analysis.

For the Bone Marrow Transplant: Children dataset, \texttt{survival\_status} was retained only for post-clustering evaluation. features representing outcome-related or post-transplant time-to-event information, including \texttt{survival\_time}, \texttt{time\_to\_aGvHD\_III\_IV}, \texttt{ANCrecovery}, and \texttt{PLTrecovery}, were removed to reduce outcome leakage and focus clustering on clinical and transplant-related characteristics.

For the Sachs protein-signaling dataset, a binary label was created by dichotomizing \texttt{PKA} at its median. The Shapiro–Wilk test in Table IV, was applied to a random sample of 5,000 observations because of the sample-size limit in the R implementation.

\subsection{t-SNE Visualizations for Three Additional Datasets} 
 
\begin{figure*}[htbp]
\centering
\captionsetup{skip=4pt}

\setlength{\tabcolsep}{2pt}
\renewcommand{\arraystretch}{0.90}

\begin{tabular}{
    >{\centering\arraybackslash}m{0.17\textwidth}
    cccccc
}

\textbf{Dataset}
&
\parbox[c]{0.15\textwidth}{\centering\small\textbf{raw data}}
&
\parbox[c]{0.15\textwidth}{\centering\small\textbf{raw-DAG}}
&
\parbox[c]{0.15\textwidth}{\centering\small\textbf{CopDAG}}
&
\parbox[c]{0.15\textwidth}{\centering\small\textbf{R-vine}} &
\parbox[c]{0.1\textwidth}{\small\textbf{Best}} &
\parbox[c]{0.1\textwidth}{\small\textbf{$2^{nd}$ Best}}
\\[7pt]
\raisebox{1cm}[0pt][0pt]{%
\parbox{0.17\textwidth}{\centering\small
\textbf{Statlog Heart\\}}}
&
\multicolumn{4}{l}{
\includegraphics[width=0.60\textwidth]
{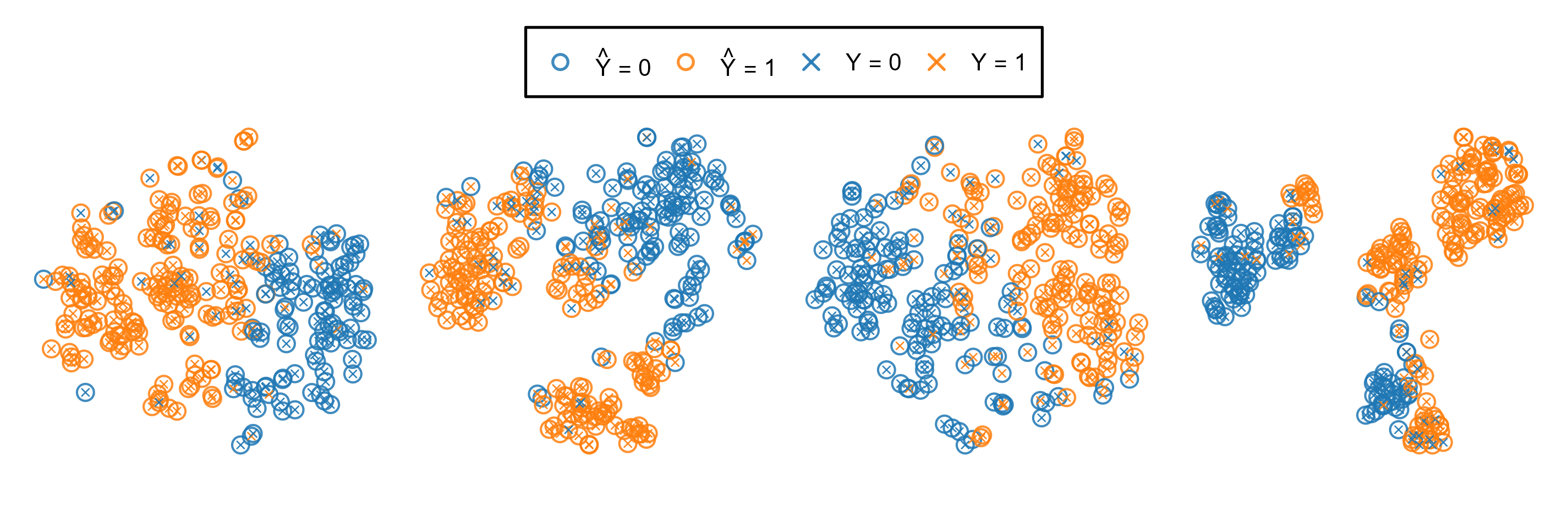}}
& \raisebox{1cm}[0pt][0pt]{%
\parbox{0.1\textwidth}{\small
\textbf{CopDAG}}} & \raisebox{1cm}[0pt][0pt]{%
\parbox{0.1\textwidth}{\small
\textbf{R-Vine}}} 

\\[-4pt]
\midrule
\raisebox{1cm}[0pt][0pt]{%
\parbox{0.17\textwidth}{\centering\small
\textbf{Bone Marrow\\Transplant}}}
&
\multicolumn{4}{l}{
\includegraphics[width=0.60\textwidth]
{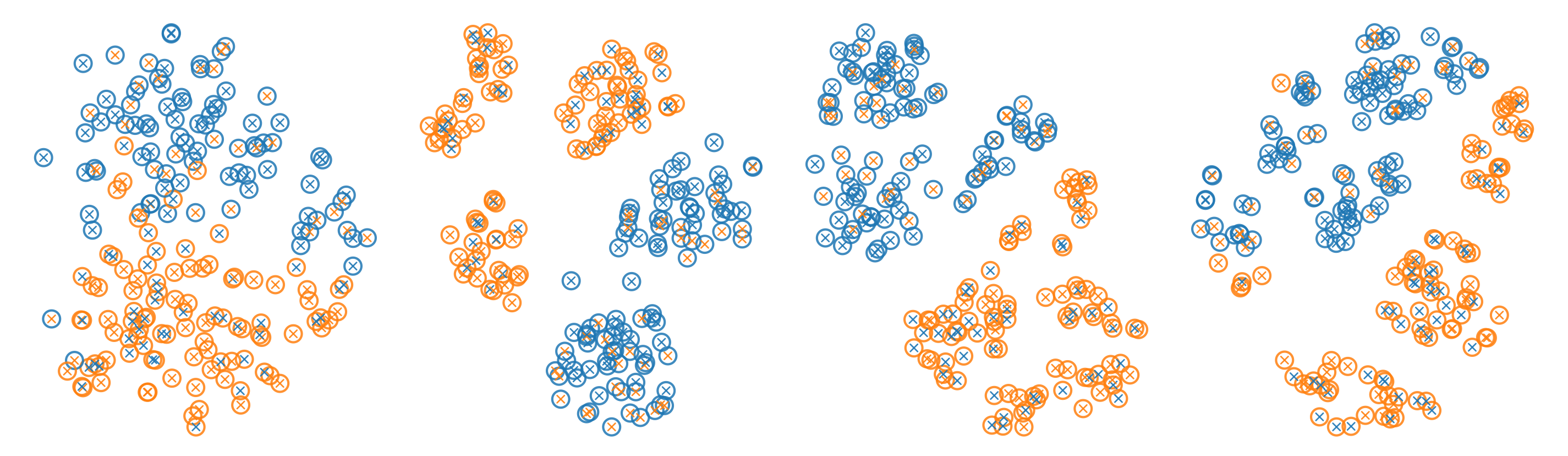}}
& \raisebox{1cm}[0pt][0pt]{%
\parbox{0.1\textwidth}{\small
\textbf{raw data}}} & \raisebox{1cm}[0pt][0pt]{%
\parbox{0.1\textwidth}{\small
\textbf{raw-DAG}}} 
\\[-4pt]
\midrule
\raisebox{1cm}[0pt][0pt]{%
\parbox{0.17\textwidth}{\centering\small
\textbf{Indian Liver\\Patient}}}
&
\multicolumn{4}{l}{
\includegraphics[width=0.60\textwidth]
{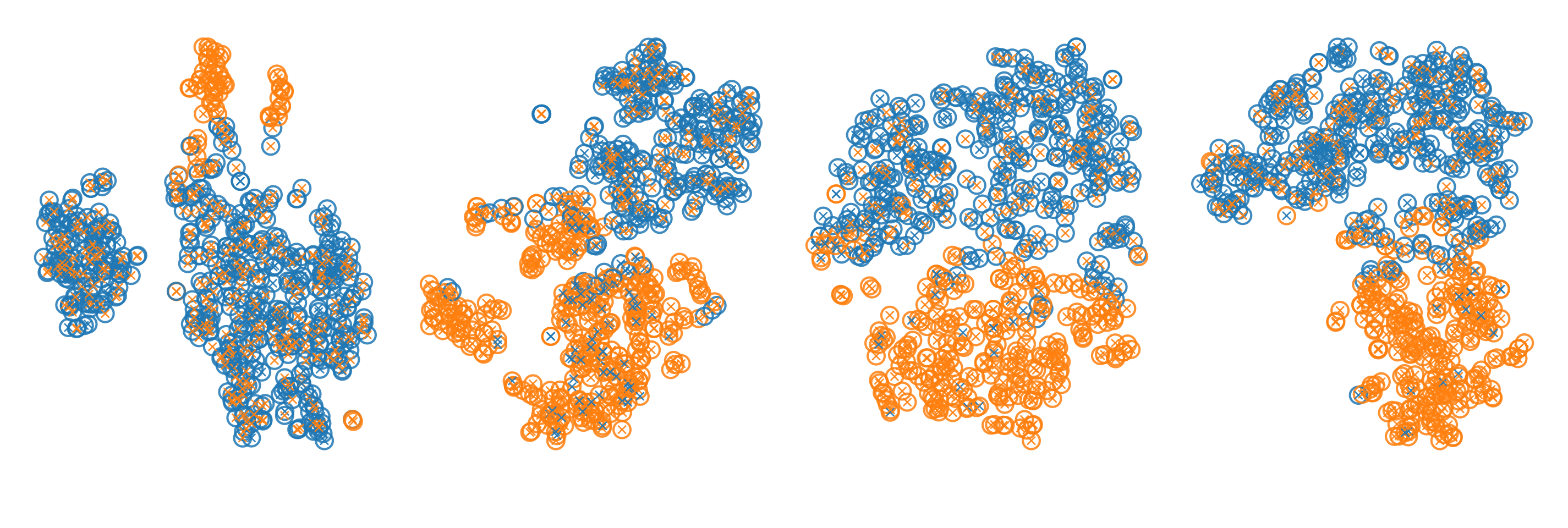}}
& \raisebox{1cm}[0pt][0pt]{%
\parbox{0.1\textwidth}{\small
\textbf{R-vine}}} & \raisebox{1cm}[0pt][0pt]{%
\parbox{0.1\textwidth}{\small
\textbf{raw data}}} 
\\[-4pt]
\end{tabular}

\caption{t-SNE visualizations using K-means clustering for three datasets}

\label{fig:tsne_kmeans_additional3_datasets}
\end{figure*}

\newpage
\subsection{t-SNE Visualizations Breast Cancer Wisconsin Dataset using K-means, HWD$^2$ and GMM Clustering} 
\begin{figure*}[htbp]
\centering
\captionsetup{skip=4pt}

\setlength{\tabcolsep}{2pt}
\renewcommand{\arraystretch}{0.90}

\begin{tabular}{
    >{\centering\arraybackslash}m{0.17\textwidth}
    cccccc
}

\parbox[c]{0.1\textwidth}{\centering\small
\textbf{Clustering\\Algorithm}}
&
\parbox[c]{0.15\textwidth}{\centering\small\textbf{raw data}}
&
\parbox[c]{0.15\textwidth}{\centering\small
\textbf{Raw-\\DAG}}
&
\parbox[c]{0.1\textwidth}{\centering\small
\textbf{Cop\\DAG}}
&
\parbox[c]{0.15\textwidth}{\centering\small\textbf{R-vine}}
&
\parbox[c]{0.1\textwidth}{\small\textbf{Best}} &
\parbox[c]{0.15\textwidth}{\small\textbf{$2^{nd}$ Best}}
\\[20pt]
\raisebox{1cm}[0pt][0pt]{%
\parbox{0.17\textwidth}{\centering\small
\textbf{K-means}}}
&
\multicolumn{4}{l}{
\includegraphics[width=0.6\textwidth]
{breast_cancer_tSNE2_true_vs_predicted_clusters2.png}}& \raisebox{1cm}[0pt][0pt]{%
\parbox{0.1\textwidth}{\small
\textbf{R-vine}}} & \raisebox{1cm}[0pt][0pt]{%
\parbox{0.15\textwidth}{\small
\textbf{CopDAG}}} 
\\[-1pt]
\midrule
\raisebox{1cm}[0pt][0pt]{%
\parbox{0.17\textwidth}{\centering\small
\textbf{HWD$^2$}}}
&
\multicolumn{4}{l}{
\includegraphics[width=0.6\textwidth]
{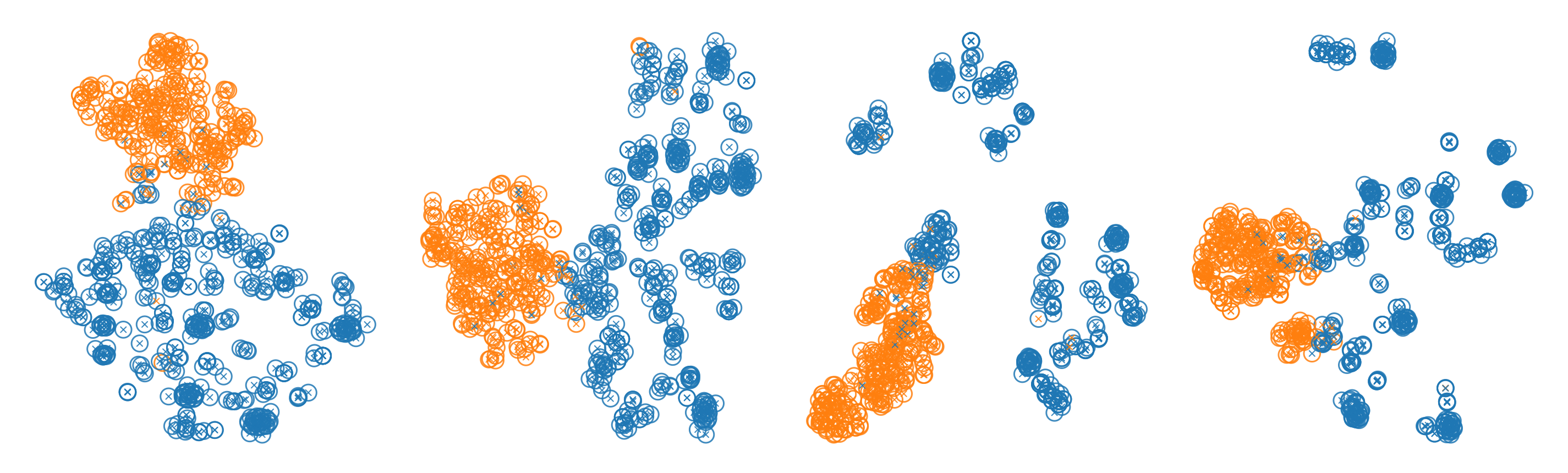}}& \raisebox{1cm}[0pt][0pt]{%
\parbox{0.1\textwidth}{\small
\textbf{raw-DAG}}} & \raisebox{1cm}[0pt][0pt]{%
\parbox{0.15\textwidth}{\small
\textbf{raw data}}} 
\\[-4pt]
\midrule
\raisebox{1cm}[0pt][0pt]{%
\parbox{0.17\textwidth}{\centering\small
\textbf{GMM}}}
&
\multicolumn{4}{l}{
\includegraphics[width=0.6\textwidth]
{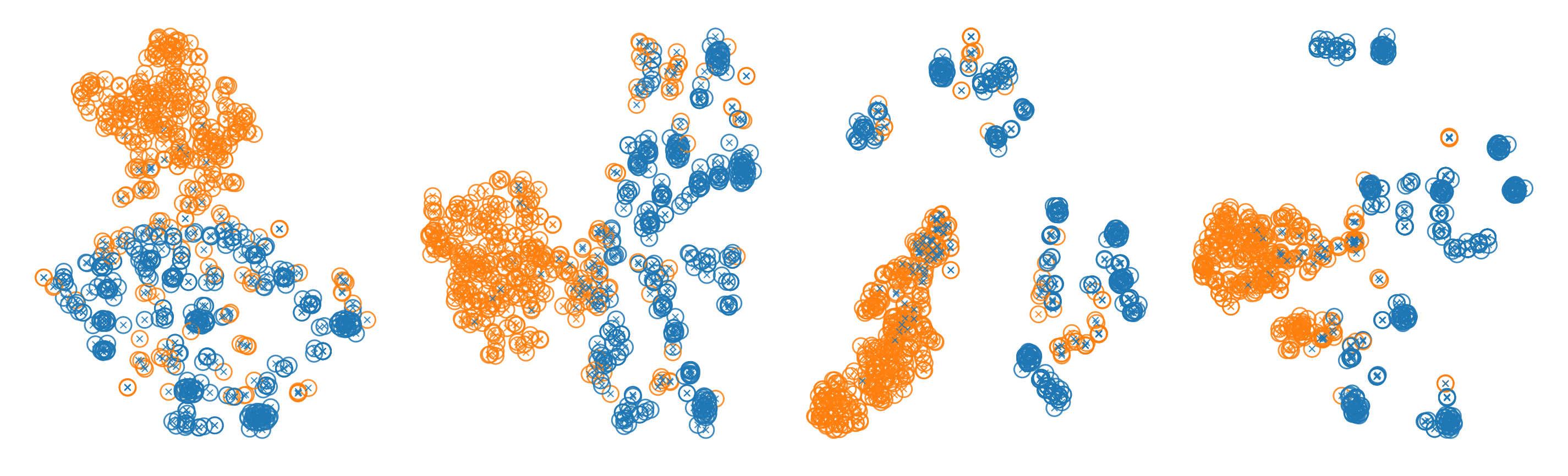}}& \raisebox{1cm}[0pt][0pt]{%
\parbox{0.1\textwidth}{\small
\textbf{R-vine}}} & \raisebox{1cm}[0pt][0pt]{%
\parbox{0.15\textwidth}{\small
\textbf{raw data}}} 
\\[-4pt]

\end{tabular}

\caption{t-SNE visualizations using K-means, HWD$^2$ and GMM clustering for Breast Cancer Wisconsin dataset. 
Open circles indicate $\hat{Y}$ and crosses indicate true labels $Y$.}

\label{fig:tsne_kmeans_HWD$^2$_gmm_breastcancer}
\end{figure*}

\newpage

\subsection{Three-dimensional t-SNE Visualizations of the K-means Clustering Results for Breast Cancer Wisconsin Dataset} 

\begin{figure}[H]
    \centering

    \includegraphics[width=0.95\textwidth]{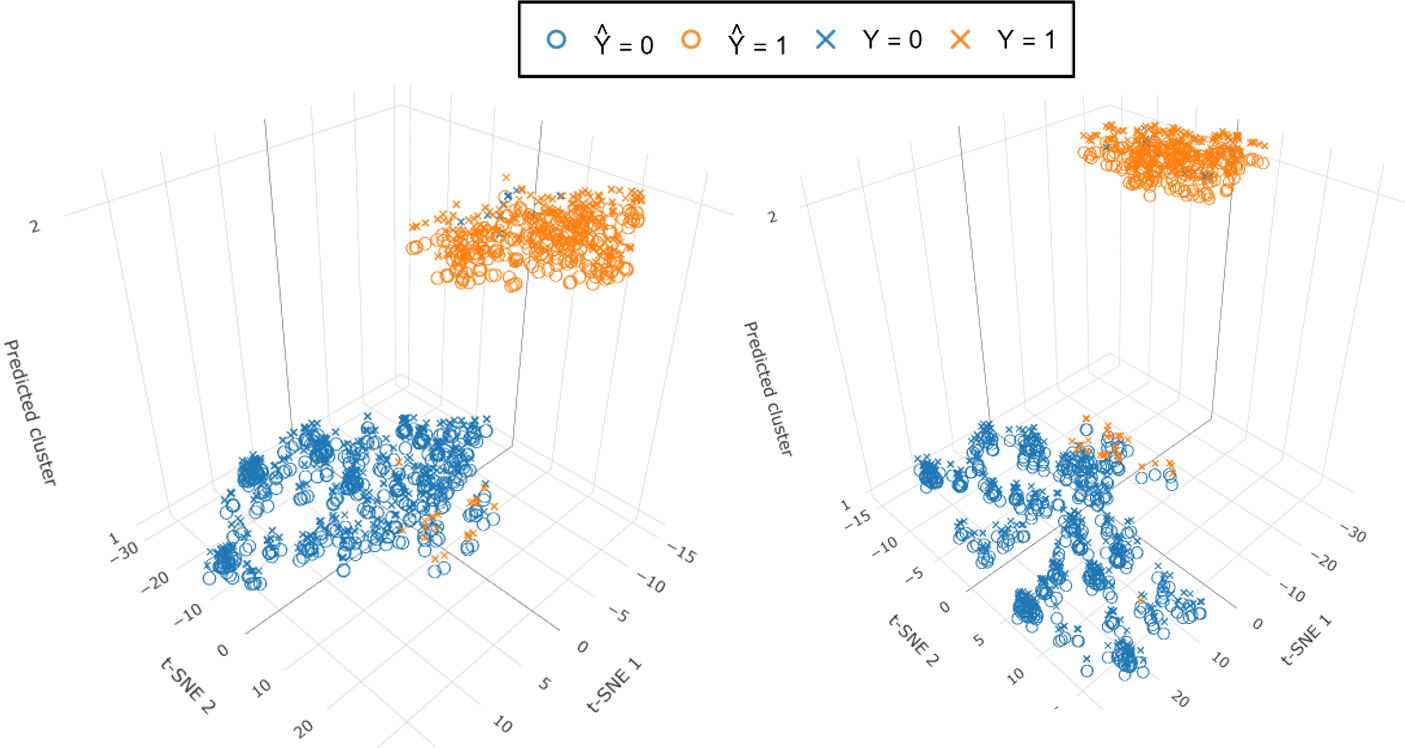}

    \vspace{-0.15cm}

    \begin{minipage}[t]{0.48\textwidth}
        \centering
        \subcaption*{\textbf{(a)} raw data}
    \end{minipage}
    \hfill
    \begin{minipage}[t]{0.48\textwidth}
        \centering
        \subcaption*{\textbf{(b)} raw-DAG}
    \end{minipage}

    \vspace{0.35cm}

    \includegraphics[width=0.95\textwidth]{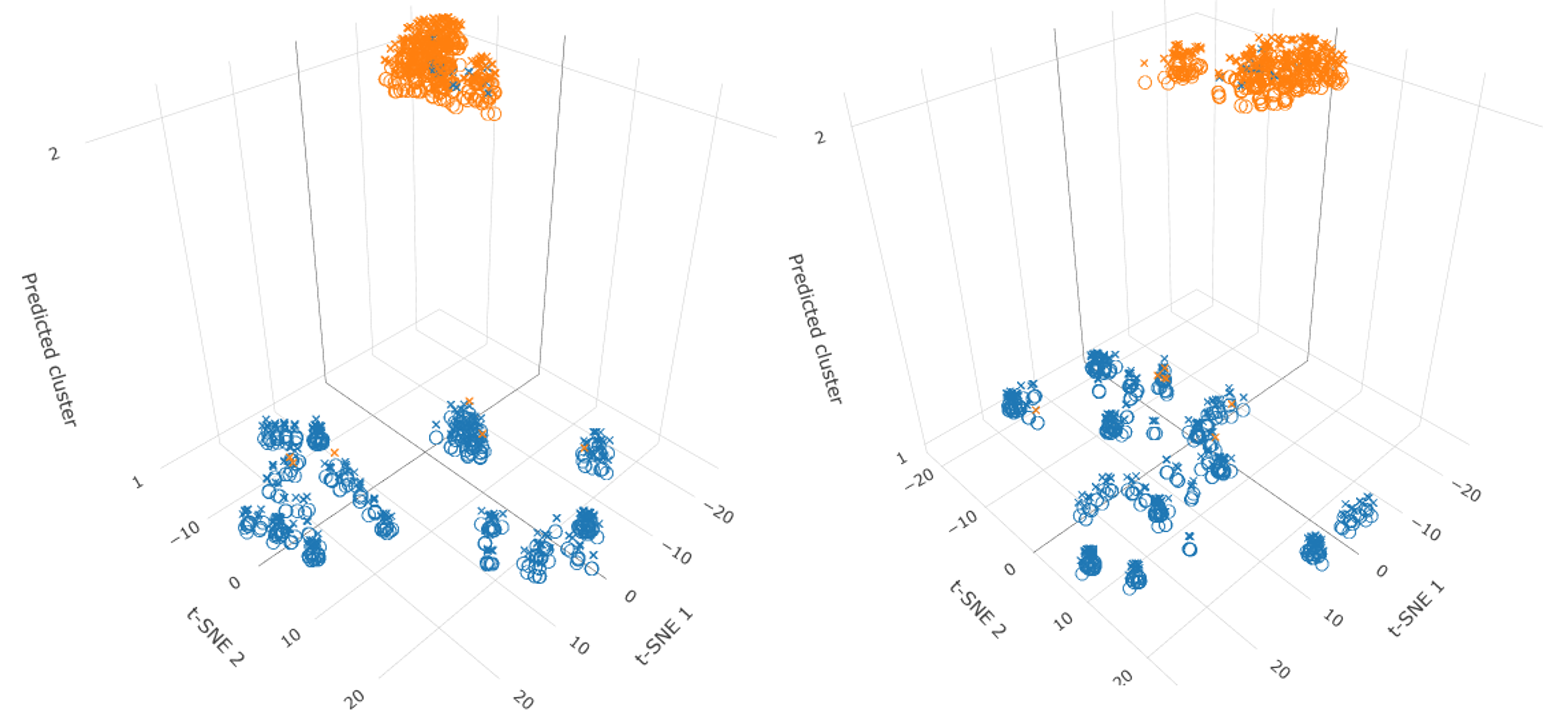}

    \vspace{-0.15cm}

    \begin{minipage}[t]{0.48\textwidth}
        \centering
        \subcaption*{\textbf{(c)} CopDAG}
    \end{minipage}
    \hfill
    \begin{minipage}[t]{0.48\textwidth}
        \centering
        \subcaption*{\textbf{(d)} R-vine}
    \end{minipage}

    \caption{Three-dimensional t-SNE visualizations of the K-means clustering results for Breast Cancer Wisconsin dataset.}
    \label{fig:kmeans_3d_tsne}
\end{figure}

\begin{table}[htbp]
\centering
\caption{$\Delta\mathrm{ARI}$ relative to the corresponding raw-data. The improved, tied, and worsened columns report the number of datasets with positive, zero, and negative changes, respectively. }
\label{tab:ari_improvement_raw_baseline}

\normalsize
\setlength{\tabcolsep}{3pt}
\resizebox{\textwidth}{!}{
\setlength{\tabcolsep}{4pt}
\renewcommand{\arraystretch}{1.1}

\begin{tabular}{llccccccc}
\toprule
\textbf{Method} &
\textbf{Clustering} &
\makecell{\textbf{Mean}\\$\boldsymbol{\Delta}$\textbf{ARI}} &
\makecell{\textbf{Median}\\$\boldsymbol{\Delta}$\textbf{ARI}} &
\textbf{Impr.} &
\textbf{Tied} &
\textbf{Wors.} &
\makecell{\textbf{Impr.}\\\textbf{$\%$}} &
\makecell{\textbf{Holm-adj}\\$\boldsymbol{p}$\textbf{-value}} \\
\midrule

R-vine
& K-means
& 0.0868
& 0.0573
& 12
& 0
& 4
& 75.0\%
& 1.0000 \\

CopDAG
& K-means
& 0.1416
& 0.0930
& 11
& 0
& 5
& 68.8\%
& 1.0000 \\

raw-DAG
& K-means
& -0.1113
& -0.0083
& 5
& 3
& 8
& 31.3\%
& 1.0000 \\

R-vine
& HWD$^2$
& 0.0923
& 0.1056
& 10
& 0
& 6
& 62.5\%
& 1.0000 \\

CopDAG
& HWD$^2$
& 0.0658
& 0.1306
& 9
& 0
& 7
& 56.3\%
& 1.0000 \\

raw-DAG
& HWD$^2$
& 0.0657
& 0.0054
& 8
& 1
& 7
& 50.0\%
& 1.0000 \\

CopDAG
& GMM
& -0.1814
& -0.0726
& 6
& 1
& 8
& 40.0\%
& 1.0000 \\

R-vine
& GMM
& -0.2467
& -0.0726
& 5
& 1
& 9
& 33.3\%
& 1.0000 \\

raw-DAG
& GMM
& -0.0078
& 0.0000
& 5
& 7
& 4
& 31.3\%
& 1.0000 \\

\bottomrule
\end{tabular}
}
\end{table}

\begin{table*}[htbp]
\centering
\caption{$\Delta\mathrm{NMI}$ relative to the corresponding raw-data. The improved, tied, and worsened columns report the number of datasets with positive, zero, and negative changes, respectively. }
\label{tab:nmi_improvement_raw_baseline}
\normalsize
\setlength{\tabcolsep}{3pt}
\resizebox{\textwidth}{!}{
\setlength{\tabcolsep}{4pt}
\renewcommand{\arraystretch}{1.1}

\begin{tabular}{llccccccc}
\toprule
\textbf{Method} &
\textbf{Clustering} &
\makecell{\textbf{Mean}\\$\boldsymbol{\Delta}$\textbf{NMI}} &
\makecell{\textbf{Median}\\$\boldsymbol{\Delta}$\textbf{NMI}} &
\textbf{Impr.} &
\textbf{Tied} &
\textbf{Wors.} &
\makecell{\textbf{Impr.}\\\textbf{$\%$}} &
\makecell{\textbf{Holm-adj}\\$\boldsymbol{p}$\textbf{-value}} \\
\midrule

R-vine
& HWD$^2$
& 0.2017
& 0.1778
& 10
& 0
& 6
& 62.5\%
& 1.0000 \\

CopDAG
& HWD$^2$
& 0.1535
& 0.1690
& 9
& 0
& 7
& 56.3\%
& 1.0000 \\

raw-DAG
& HWD$^2$
& 0.0232
& 0.0230
& 9
& 1
& 6
& 56.3\%
& 1.0000 \\

R-vine
& K-means
& -0.0305
& 0.0551
& 8
& 0
& 8
& 50.0\%
& 1.0000 \\

CopDAG
& K-means
& -0.0418
& -0.0137
& 8
& 0
& 8
& 50.0\%
& 1.0000 \\

raw-DAG
& K-means
& -0.2035
& -0.0636
& 3
& 3
& 4
& 18.8\%
& 0.4533 \\

CopDAG
& GMM
& -0.0880
& 0.0174
& 8
& 1
& 6
& 53.3\%
& 1.0000 \\

R-vine
& GMM
& -0.1305
& 0.0057
& 8
& 1
& 6
& 53.3\%
& 1.0000 \\

raw-DAG
& GMM
& 0.0051
& 0.0000
& 4
& 7
& 5
& 25.0\%
& 1.0000 \\

\bottomrule
\end{tabular}
}
\end{table*}

\end{document}